\documentclass{article}

\PassOptionsToPackage{numbers,compress}{natbib}

 \usepackage[preprint]{neurips_2026}

\usepackage[utf8]{inputenc} 
\usepackage[T1]{fontenc}    
\usepackage{hyperref}       
\usepackage{url}            
\usepackage{booktabs}       
\usepackage{amsfonts}       
\usepackage{nicefrac}       
\usepackage{microtype}      
\usepackage{xcolor}         

\usepackage{multirow}
\usepackage{amsmath}
\usepackage{graphicx}
\usepackage{makecell}
\usepackage{pifont}
\usepackage{colortbl}
\usepackage{array}

\usepackage{amsthm}

\newcolumntype{C}{>{\centering\arraybackslash}p{1.1cm}}

\newcommand{\ve}[1]{\mathbf{#1}}
\newcommand{\matr}[1]{\mathrm{#1}}
\newcommand{\Std}[1]{#1}

\title{SlimSpec: Low-Rank Draft LM-Head for Accelerated Speculative Decoding}

\author{
Anton~Plaksin\thanks{Nebius, Amsterdam, Netherlands. \\Correspondence to: \texttt{astrlrd@nebius.com}} \hspace{1em}
Sergei~Krutikov\footnotemark[1] \hspace{1em}
Sergei~Skvortsov\footnotemark[1] \hspace{1em}
Alexander~Samarin\footnotemark[1]
}

\begin{document}

\maketitle

\begin{abstract}
Speculative decoding speeds up autoregressive generation in Large Language Models (LLMs) through a two-step procedure, where a lightweight draft model proposes tokens which the target model then verifies in a single forward pass.
Although the drafter network is small in modern architectures, its LM-head still performs projection to a large vocabulary, becoming one of the major computational bottlenecks.
In prior work this issue has been predominantly addressed via static or dynamic vocabulary truncation.
Yet mitigating the bottleneck, these methods bring in extra complexity, such as special vocabulary curation, sophisticated inference-time logic or modifications of the training setup.
In this paper, we propose \textbf{SlimSpec}, a low-rank parameterization of the drafter's LM-head that compresses the inner representation rather than the output, preserving full vocabulary support.
We evaluate our method with EAGLE-3 drafter across three target models and diverse benchmarks in both latency- and throughput-bound inference regimes.
SlimSpec achieves $4\text{-}5\times$ acceleration over the standard LM-head architecture while maintaining a competitive acceptance length, surpassing existing methods by up to $8\text{-}9\%$ of the end-to-end speedup.
Our method requires minimal adjustments of training and inference pipelines.
Combined with the aforementioned speedup improvements, it makes SlimSpec a strong alternative across wide variety of draft LM-head architectures.
\end{abstract}

\section{Introduction}\label{sec:intro}

In the past years, Large Language Models (LLMs) have achieved strong performance across a wide range of tasks, but their autoregressive nature remains computationally inefficient at inference due to sequential token generation.
As a result, latency and serving costs have become significant challenges for practical deployment. 
A central direction for mitigating these costs is \emph{speculative decoding}~\citep{leviathan2023fast,chen2023accelerating} that employs a lightweight draft model to propose multiple consecutive tokens which the target model then verifies in parallel.
This procedure accelerates generation by sampling multiple tokens per speculative round on average, while preserving the output distribution of the target model.

Since its introduction, speculative decoding has evolved into a broad family of methods.
Early approaches used standalone drafters, including pretrained small language models from the same model family or simple n-gram drafters that derive proposals from corpus statistics or the prompt context~\citep{he2023rest,fu2024lookahead}.
More recent methods, such as MEDUSA, Hydra and the EAGLE family~\citep{cai2024medusa,ankner2024hydra,li2024eagle,li2024eagle2,li2025eagle3}, integrate a lightweight drafter module into the target model directly, building it upon extracted hidden representations.

This design has become a commonly used approach due to lower overhead and higher acceptance quality, resulting into substantial improvements in the end-to-end speedups.
One of its major bottlenecks, which limits further speedup advances, is computation of the draft token logits~\citep{zhao2025frspec,williams2026specvocab}.
Although the drafter backbone can be small, its LM-head has to produce logits over the whole target model vocabulary whose size in modern LLMs often exceeds the order of $10^5$.
This requires a large output projection at every drafted position, making the LM-head a natural computational bottleneck.


Existing methods mainly mitigate the aforementioned issue by shrinking the active vocabulary, either statically~\citep{zhao2025frspec,goel2025vocabtrim,shoham2026balancing} or dynamically~\citep{williams2026specvocab,weng2025coral,zhang2025dynaspec}, thereby reducing the output projection along the vocabulary dimension.
While being effective, these methods introduce additional complexity, such as vocabulary curation, token-index bookkeeping, inference-time routing or top-$k$ selection.

\begin{figure}[t]
  \centering
  \includegraphics[width=\linewidth]{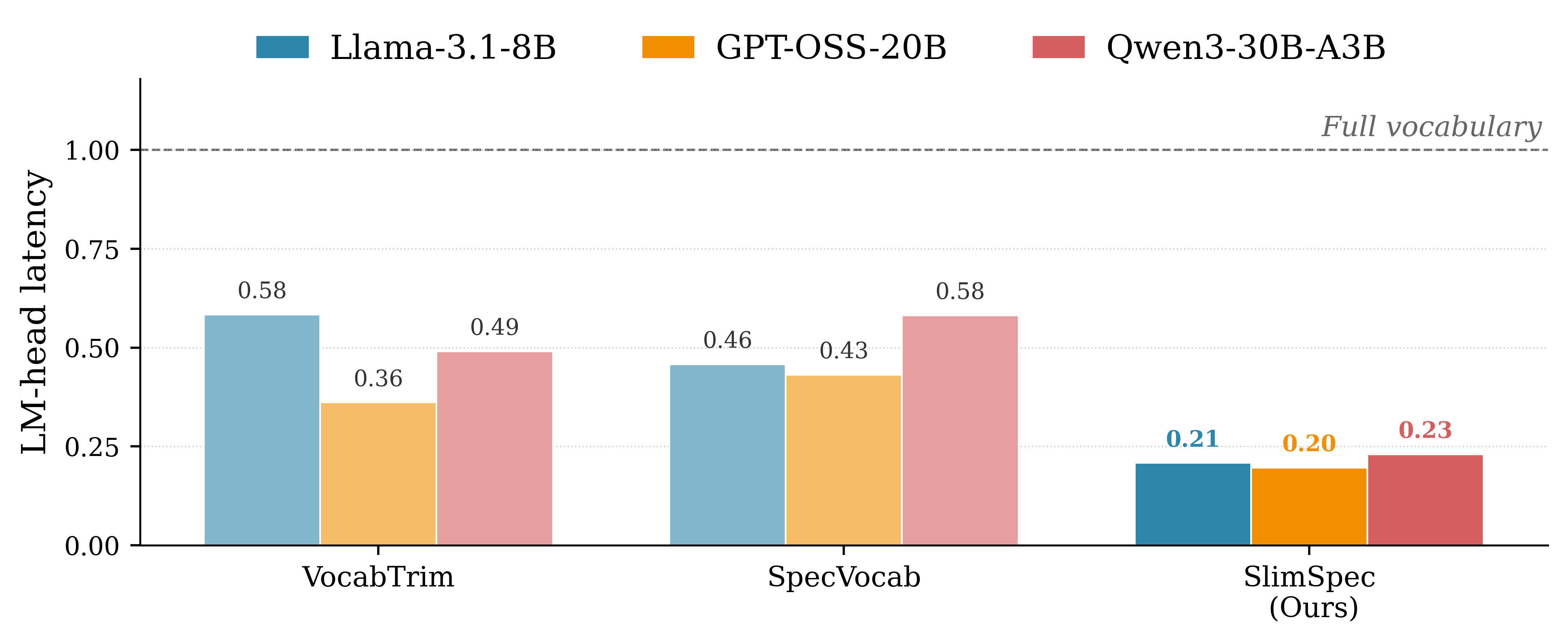}
  \caption{
  Relative LM-head GPU time $T_{\mathrm{head}}$ for batch size $1$ across models; lower is better.
  The underlying $T_{\mathrm{head}}$ values are normalized with respect to the full vocabulary baseline, set to $1.0$. 
  VocabTrim reduces the draft vocabulary to $64$K tokens.
  For SpecVocab and SlimSpec the low rank is set to $r=d/8$, where $d$ is target model hidden size.
  Both VocabTrim and SpecVocab can reduce LM-head latency by only about \(60\%\), while SlimSpec achieves an approximately \(4\text{-}5\times\) reduction.
  }\label{fig:main_results}
\end{figure}

In this paper, we explore a different direction of addressing the LM-head bottleneck.
Instead of reducing the set of candidate tokens, we compress the hidden representation being used for logits prediction.
Our approach preserves the full vocabulary of the target model, keeps computations dense and needs minimal changes in both training and inference pipelines.
Our contributions are as follows:
\begin{itemize}
    \item We propose \textbf{SlimSpec}, a drop-in low-rank LM-head architecture for speculative drafters.
    It replaces the standard LM-head with a factorized projection that compresses the hidden representation rather than the output vocabulary.
    We rigorously show that our approach is free of theoretical drawbacks inherent to vocabulary reduction methods, including hard ceiling on the acceptance rate and train-test mismatch of the target model distribution.


    \item We derive an acceptance-cost framework that reveals when LM-head acceleration translates into the end-to-end speedup.
    Our analysis establishes the relation between computational speedup and acceptance preservation, which helps finding a reasonable trade-off for better overall performance and provides the ground for comparing different LM-head designs.
    
    \item We validate this analysis in a production-like inference setup using EAGLE-3 drafter across three target models (Llama3.1-8B, GPT-OSS-20B, Qwen3-30B-A3B), diverse benchmarks, different decoding temperatures and serving regimes.
    Under identical training pipelines and serving configurations, SlimSpec maintains acceptance length close to the full-vocabulary baseline while reducing LM-head latency by approximately $4\text{-}5$ times as shown in Figure~\ref{fig:main_results}, surpassing other evaluated methods by $8\text{-}9\%$ in terms of the end-to-end speedup.
    

\end{itemize}


\section{Related Work}\label{sec:related_work}

\begin{table}[t]
\centering
\caption{\textbf{Complexity and design parameters of LM-head acceleration methods.}
\textit{No Vocabulary Pruning} indicates whether the full vocabulary is preserved.
\textit{No Top-$k$ Overhead} indicates whether the LM head needs top-$k$ token selection at inference.
\textit{Hyperparameters} lists configuration choices required by each method, beyond standard drafter training settings.
\textit{LM Head Complexity} gives asymptotic FLOPs required to compute LM head forward pass per drafted token, where $V = |\mathcal{V}|$ is the full vocabulary size and $d$ is the drafter hidden-state dimension.}
\label{tab:complexity}
\setlength{\tabcolsep}{4pt}
\renewcommand{\arraystretch}{1.45}
\begin{tabular}{@{}l c c l l@{}}
\toprule
\textbf{Method}
& \textbf{\makecell{No Vocabulary\\Pruning}}
& \textbf{\makecell{No Top-$k$\\Overhead}}
& \textbf{Hyperparameters}
& \textbf{\makecell{LM Head\\Complexity}} \\
\midrule
Full Vocab
  & \ding{51}
  & \ding{51}
  & ---
  & $\mathcal{O}(Vd)$ \\
\addlinespace[6pt]

\makecell[l]{
FR-Spec~\cite{zhao2025frspec}\\
VocabTrim~\cite{goel2025vocabtrim}\\
BCL~\cite{shoham2026balancing}
}
  & \ding{55}
  & \ding{51}
  & $V_{\mathrm{tr}}$: truncated vocab size
  & $\mathcal{O}(V_{\mathrm{tr}}d)$ \\
\addlinespace[6pt]

CORAL~\cite{weng2025coral}
  & \ding{55}
  & \ding{55}
  & \makecell[l]{$N$: \# groups\\ $k$: \# activated groups}
  & $\mathcal{O}\!\left(d^2 + Nd + k\frac{Vd}{N}\right)$ \\
\addlinespace[6pt]

DynaSpec~\cite{zhang2025dynaspec}
  & \ding{55}
  & \ding{55}
  & \makecell[l]{$B$: shortlist size\\ $r$: router interm. size\\ $M$: \# clusters}
  & $\mathcal{O}(2rd + rM + Bd)$ \\
\addlinespace[6pt]

SpecVocab~\cite{williams2026specvocab}
  & \ding{51}
  & \ding{55}
  & \makecell[l]{$r$: router low rank\\ $k$: \# selected tokens}
  & $\mathcal{O}(rd + Vr + kd)$ \\
\midrule

\rowcolor{gray!15}
\textbf{SlimSpec (ours)}
  & \ding{51}
  & \ding{51}
  & $r$: low rank
  & $\boldsymbol{\mathcal{O}(rd + Vr)}$ \\
\bottomrule
\end{tabular}
\end{table}

Recent work has increasingly focused on reducing the complexity of the drafter LM-head in speculative decoding. We classify existing methods into two families.

The first family reduces the drafter-side cost through \emph{static vocabulary truncation}.
FR-Spec~\citep{zhao2025frspec} and VocabTrim~\citep{goel2025vocabtrim} share the same core idea: they truncate draft prediction to a smaller token set, thereby reducing the cost of the drafter LM-head.
The main difference lies in the source of the frequency statistics used to choose this truncated vocabulary.
VocabTrim ranks tokens by their frequency in target-model sampled generations whereas FR-Spec ranks tokens by their occurrence frequency in a general-purpose text corpus.
More recently, BCL~\citep{shoham2026balancing} studied static truncation as an optimization problem that balances token coverage against draft-side latency through the choice of vocabulary size.
Unlike VocabTrim and FR-Spec, BCL explicitly trains the drafter with the found optimal vocabulary, thereby aligning training and inference.
Similarly, \citep{samarin2026lk} also train draft models with truncated output vocabularies, although their main focus is the training objective rather than vocabulary selection.

The advantage of this family is architectural simplicity, but its limitation is inherent to vocabulary truncation. 
All tokens outside this vocabulary are assigned zero probability and can never be proposed by the drafter, which typically reduces acceptance quality. 
Recent work~\citep{timor2025out} mitigates this limitation by redistributing drafter probability mass toward target tokens outside the truncated vocabulary.
However, this primarily serves as an acceptance-recovery mechanism for pruned vocabularies, rather than a method for making the LM-head computation itself cheaper.

The second family of methods perform \emph{dynamic selection} of the active vocabulary subset.
CORAL~\citep{weng2025coral} and DynaSpec~\citep{zhang2025dynaspec} both rely on a predefined partition of the vocabulary into small disjoint subsets and add a routing mechanism that selects a few active subsets for each context.
The logits are then computed only over these selected subsets, reducing LM-head cost while allowing the active support to depend on the current context.
SpecVocab~\citep{williams2026specvocab} follows a related routing-based approach but avoids predefined expert sub-vocabularies.
Instead of routing to these fixed partitions, it uses a learned low-rank router to predict a context-dependent token subset directly. 

Compared with static vocabulary truncation, these methods can improve a trade-off between the speedup and acceptance quality by adapting the candidate vocabulary to the context.
However, this flexibility comes at the cost of a more sophisticated design and implementation.
As shown in Table~\ref{tab:complexity}, these methods introduce more hyperparameters and require an explicit top-\(k\)-style selection step before the final logit computation.
The latter can become a noticeable bottleneck on GPUs because it involves such operations as global ranking, partial sorting, irregular indexing and gathering a context-dependent subset of weights, which are less efficient than dense matrix multiplication.

We also note a broader line of work that compresses the LM-head via low-rank architectural factorization in standard language modeling~\citep{grave2017adaptive,chen2018groupreduce,baevski2019adaptive,lioutas2019distilled,hrinchuk2020tensorized,lan2020albert,maalouf2021deep}.
These methods are not specific to speculative decoding and operate on a single model, so they are not applicable to the drafter-target setup considered here.

\section{Performance Model for Draft LM-Head Acceleration}\label{sec:performance_model}

In this section, we analyze the throughput structure of speculative decoding and quantify the contribution of the drafter LM-head to the drafting cost.
We also derive an acceptance–cost trade-off that governs how reducing this cost translates into end-to-end speedup.

\subsection{Throughput structure of speculative decoding}
Let $n$ be the maximum number of draft tokens proposed per speculative step.
Following the standard convention~\citep{leviathan2023fast}, we measure acceptance quality by the \textit{average acceptance length} defined as
\begin{equation}\label{eq:tau}
\tau = n \times \frac{\#\text{accepted tokens}}{\#\text{drafted tokens}} + 1 .
\end{equation}
Here, the first term estimates the average number of accepted draft tokens per speculative step, while the $1$ accounts for the bonus token sampled from the target-model distribution after verification.

Let \(T_{\mathrm{draft}}\) be the wall-clock time required to generate the draft tokens, \(T_{\mathrm{verify}}\) be the wall-clock time of the target-model verification pass, and \(T_{\mathrm{overhead}}\) be the pipeline overhead, including scheduling, synchronization, and cache management. 
Then the decoding throughput can be written~\citep{leviathan2023fast,weng2025coral,zhang2025dynaspec} as the average number of \emph{tokens per second}
\begin{equation}\label{eq:tps}
    \mathrm{TPS} =
    \frac{\tau}
    {T_{\mathrm{overhead}} + T_{\mathrm{verify}} + T_{\mathrm{draft}}}.
\end{equation}

\begin{figure}[t]
    \centering
    \includegraphics[width=\linewidth]{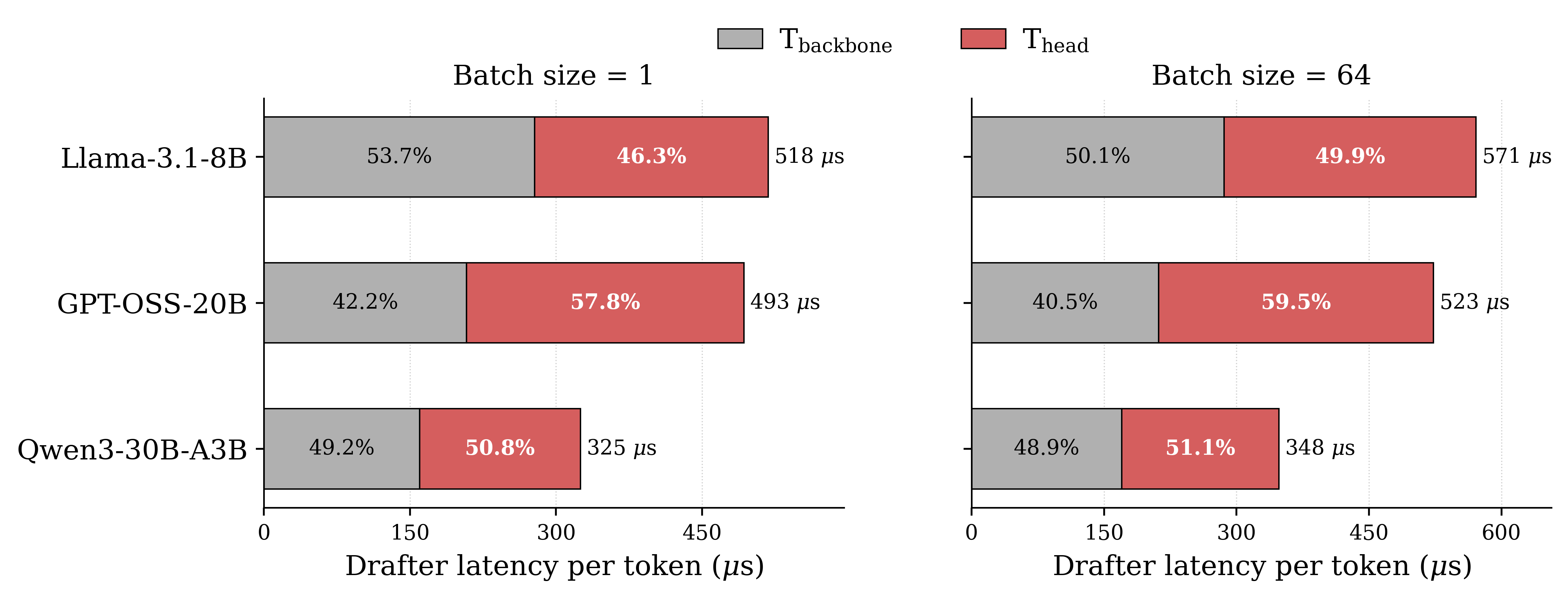}
    \caption{
    Drafter latency decomposition at batch sizes $1$ and $64$ for EAGLE-3 with full-vocabulary projection. Each horizontal bar breaks down $T_{\mathrm{draft}}$ into main block computation time $T_{\mathrm{backbone}}$ and LM-head computation time $T_{\mathrm{head}}$ for each target model, measured in microseconds.
    }\label{fig:draft_cost}
\end{figure}

We further focus on modern auxiliary-head drafter architectures, such as MEDUSA, Hydra and the EAGLE family.
Despite differences in their specific designs, their draft-side computation naturally separates into backbone components that produce draft hidden states and a final LM-head projection that maps these states to vocabulary logits. 
To isolate the LM-head cost \(T_{\mathrm{head}}\) from the remaining computations \(T_{\mathrm{backbone}}\), we decompose
\begin{equation}\label{eq:draft_time}
T_{\mathrm{draft}}
=
T_{\mathrm{backbone}}
+
T_{\mathrm{head}}.
\end{equation}
The main observation motivating our work is that \(T_{\mathrm{head}}\) accounts for roughly \(\mathbf{45\% \text{-} 60\%}\) of \(T_{\mathrm{draft}}\), depending on the target model and inference regime (see Figure~\ref{fig:draft_cost}).
The reason is structural: the draft model must remain lightweight in order to provide a speedup, yet at each drafted position it still has to produce a distribution over the entire target model vocabulary $\mathcal{V}$.
The standard full-vocabulary projection has complexity \(\mathcal{O}(Vd)\),
where $V = |\mathcal{V}|$ is the vocabulary size and $d$ is the hidden state dimension of the drafter.
In modern LLMs, this corresponds to \textbf{hundreds of millions of operations per drafted token} and therefore makes the drafter LM-head a natural computational bottleneck.

\subsection{Acceptance–cost trade-off}\label{subsec:speedup_improvement}
Reducing the LM-head cost is only useful if it translates into end-to-end speedup.
For a method $M$\footnote{
By the method $M$ we mean a particular design of the draft LM head, with $M=\mathrm{Full}$ being a standard LM-head projection to the full target vocabulary.
Accordingly, $\tau_M$ and $T_{\mathrm{head}}^M$ correspond to the respective quantities measured using identical setups except different LM heads.
}
with mean acceptance length $\tau_M$ and head latency $T_{\mathrm{head}}^M$, define
\[
\rho_\tau \;=\; \frac{\tau_M}{\tau_{\mathrm{Full}}}, \qquad \nu \;=\; \frac{T_{\mathrm{head}}^M}{T_{\mathrm{head}}^{\mathrm{Full}}}, \qquad \kappa \;=\; \frac{T_{\mathrm{head}}^{\mathrm{Full}}}{T_{\mathrm{non\text{-}head}}},
\]
where $T_{\mathrm{non\text{-}head}} = T_{\mathrm{overhead}} + T_{\mathrm{verify}} + T_{\mathrm{backbone}}$.
Here $\rho_\tau$ measures acceptance preservation, $\nu \in (0, 1]$ determines LM-head acceleration, and $\kappa$ quantifies how much the LM-head dominates the rest of the speculation pipeline. 
Using the throughput formula~\eqref{eq:tps}, the end-to-end speedup of $M$ relative to the full-vocabulary baseline is
\begin{equation}
\rho_{\mathrm{TPS}}(\nu, \rho_\tau; \kappa) \;=\; \rho_\tau \cdot \frac{1 + \kappa}{1 + \nu\kappa}.
\label{eq:speedup}
\end{equation}

Equation~\eqref{eq:speedup} defines a family of speedup level curves on the $(\nu, \rho_\tau)$ plane, parameterized by $\kappa$.
A method with parameters $(\nu, \rho_\tau)$ provides end-to-end speedup improvement over the full-vocabulary baseline if and only if
\begin{equation}
\rho_\tau \;>\; \frac{1 + \nu\kappa}{1 + \kappa}.
\end{equation}

The right-hand side defines the minimum acceptance ratio which method $M$ must preserve in order to convert its LM-head savings into end-to-end gains. 
If the LM-head accounts for a small fraction of pipeline costs, $\kappa \to 0$, the threshold approaches $1$ and any acceptance loss is fatal. 
When the LM-head dominates, the threshold becomes smaller and more substantial acceptance loss is tolerated.

Parameter $\kappa$ is not a property of the drafter alone but of the full deployment configuration. 
Larger or deeper target models likewise increase $T_{\mathrm{verify}}$ and lower $\kappa$. 
Batch size can switch individual pipeline components between memory- and compute-bound regimes, shifting the relative weight of $T^{\mathrm{Full}}_{\mathrm{head}}$ and $T_{\mathrm{non\text{-}head}}$.
Sampling temperature also plays a role: stochastic decoding requires a softmax over the whole vocabulary, increasing $T_{\mathrm{overhead}}$ and thereby reducing $\kappa$ relative to greedy decoding.
Other factors include sampling tokens from the residual distribution, computing acceptance probabilities and performing stochastic rejection sampling itself.
Finally, since the standard LM-head scales as $\mathcal{O}(Vd)$ while the rest of the drafter scales with $d^2$, target models with larger vocabularies (relative to~$d$) push $\kappa$ upward.

\section{SlimSpec}\label{sec:slimspec}

The framework introduced in Section~\ref{subsec:speedup_improvement} establishes the condition when LM-head acceleration translates to end-to-end speedup improvements. 
LM head must achieve a sufficiently low latency factor $\nu$ without sacrificing too much acceptance ratio $\rho_\tau$, with the exact trade-off governed by the parameter $\kappa$.
In this section, we introduce SlimSpec, a lightweight LM-head architecture for speculative decoding, whose design is driven by the aforementioned principles.


\subsection{LM-head architecture}
Let $\ve{h} \in \mathbb{R}^d$ denote the hidden representation produced by the draft model backbone, and $\ve{z}\in\mathbb{R}^{V}$ be the corresponding logits.
The standard full-vocabulary projection is
\[
\ve{z} = \matr{W}_{\mathrm{full}}\, \ve{h},\quad \matr{W}_{\mathrm{full}} \in \mathbb{R}^{V \times d}.
\]
SlimSpec replaces it with the low-rank factorization
\[
\ve{z} = \matr{W}_{\mathrm{up}} \matr{W}_{\mathrm{down}} \ve{h},\quad \matr{W}_{\mathrm{down}} \in \mathbb{R}^{r \times d},\quad \matr{W}_{\mathrm{up}} \in \mathbb{R}^{V \times r},
\]
where $r < d$ is the chosen rank\footnote{
Although our parameterization might resemble a well-known LoRA-style adapters, its role is different.
LoRA adds a low-rank update to an existing full-rank matrix and therefore preserves the full-rank path.
In contrast, SlimSpec completely removes the full-rank LM-head and replaces it by the low-rank representation.
}.
The full target vocabulary is preserved through $\matr{W}_{\mathrm{up}}$, while LM-head computational cost reduces from $\mathcal{O}(Vd)$ to $\mathcal{O}(rd+Vr)$.
Since $V\gg d$ in modern LLMs, the cost reduction (in FLOPs) is approximately linear in $r$:
\[
\frac{rd+Vr}{Vd}=\frac{r}{d}\cdot\left(1+\frac{d}{V}\right)\approx\frac{r}{d}.
\]

Conceptually, the vocabulary is not trimmed - instead, the hidden representation used for logits prediction is compressed.
This is the main distinction between SlimSpec and vocabulary-truncation approaches: it keeps all \(V\) token logits available while generating them via a thinner representation.


The rank \(r\) is the only architectural hyperparameter of SlimSpec, which positively distinguishes it from dynamic vocabulary truncation methods.
It controls both the width of the compressed hidden state and the computational cost of the head.
In practice, useful ranks are fractions of the drafter hidden dimension, such as \(d/4\), \(d/8\) or \(d/16\).
We further study this trade-off empirically in Section~\ref{sec:eval_results}.

\subsection{Advantages over vocabulary truncation}\label{subsec:slimspec_advantages}
The central design decision in SlimSpec is compressing the hidden representation rather than restricting the output support.
We argue here that this choice is structurally superior to vocabulary truncation due to two key properties: output support preservation and a train-test consistency.

\paragraph{Acceptance upper bound}
Let \(p\) and \(q\) be the target and draft distributions respectively at a given draft position.
The acceptance rate for this position is governed by the distributional overlap
\[
\alpha = \sum_{v \in \mathcal{V}} \min(p(v), q(v)).
\]
If the drafter is restricted to a truncated vocabulary \(\mathcal{V}_{\mathrm{tr}}\subset \mathcal{V}\), then \(q(v)=0\) for \(v \notin \mathcal{V}_{\mathrm{tr}}\), which implies
\[
\alpha
\le
\sum_{v \in \mathcal{V}_{\mathrm{tr}}} p(v)
\]
for \emph{any} draft distribution $q$\footnote{
Under greedy decoding ($T = 0$), target distribution becomes a point-mass on $v^* = \arg\max_{v\in \mathcal{V}}p(v)$, so the bound reduces to $\alpha \leq \mathbf{1}[v^* \in \mathcal{V}_{\mathrm{tr}}]$ and acceptance collapses to $0$ whenever $v^* \notin \mathcal{V}_{\mathrm{tr}}$.
}.
This bound holds for all drafters with the truncated vocabulary, regardless of training quality, parameter count, routing scheme or alike.
SlimSpec is not subject to this bound.

\paragraph{Train-test mismatch under KL training}
A more subtle problem arises when vocabulary truncation is combined with a forward Kullback-Leibler divergence loss during training of the drafter. 
Let $\ve{z}_p\in\mathbb{R}^{V}$ be the logits of the target model, so \(p=\operatorname{softmax}(\ve{z}_p)\).
When the drafter's LM-head is restricted to \(\mathcal{V}_{\mathrm{tr}}\), the KL divergence becomes infinite since \(q(v)=0\), \(\forall v \notin \mathcal{V}_{\mathrm{tr}}\), whilst \(p(v) > 0\) for all finite logits.
In practice, this inconsistency is resolved by redefining the target as
\[
\tilde{p}(v) = \operatorname{softmax}(\ve{m} \odot \ve{z}_p),
\]
where mask \(\ve{m}\) sets logits $\ve{z}_p$ to \(-\infty\) for the tokens outside \(\mathcal{V_{\mathrm{tr}}}\)~\citep{samarin2026lk}.

This introduces a discrepancy between target distributions being used in the training objective and in the test-time acceptance logic. 
At inference, the drafter is verified against the full target distribution \(p\) whereas during training it only sees a truncated and re-normalized distribution \(\tilde{p}\).
Therefore, KL-based training is likely to produce overconfident draft probabilities on $\mathcal{V_{\mathrm{tr}}}$ at large scale, which may harm acceptance rates since overshooting test-time target probabilities reduces acceptance probabilities.
 
\paragraph{Simplicity}
Unlike vocabulary truncated methods, SlimSpec does not require complex data preprocessing to compute token statistics or storing and manipulating token index mappings.
It only consists of regular dense matrix multiplications, avoiding poorly scalable routing or top-$k$-selection logic.
SlimSpec requires small modification of the LM head and can be seamlessly plugged into any existing drafter without altering its backbone or training pipeline.
This makes our method substantially easier to implement than competing approaches.

\section{Experimental Settings}

\subsection{Methods}\label{subsec:methods}
We compare SlimSpec, as it is introduced in Section~\ref{sec:slimspec}, against three groups of baselines described below.
As the default, we use a standard approach that performs a linear projection to the full target vocabulary.
We further refer to this baseline as Full Vocab.

For static vocabulary truncation, we consider two post-training baselines, VocabTrim and FR-Spec. 
Following the methodology of the original papers, we select a token subset by ranking vocabulary according to the frequency statistics collected on the calibration dataset.
In VocabTrim it is simply the training dataset whereas in FR-Spec it is a general-purpose SlimPajama-627B~\citep{soboleva2023slimpajama}.

As a training-aware baseline, we consider BCL which performs vocabulary truncation according to the optimal coverage-latency trade-off.
We also report VocabTrim-T which trains the drafter using the same truncated vocabulary as VocabTrim.

For dynamic vocabulary truncation, we consider SpecVocab due to its simplicity and strong performance.
We implement this method following the original paper and report results for several values of the router rank $r$.

\subsection{Training configuration}\label{subsec:training_config}
We conduct experiments across three target models: Llama-3.1-8B-Instruct~\citep{grattafiori2024llama3}, GPT-OSS-20B~\citep{agarwal2025gptoss}, and Qwen3-30B-A3B-Instruct-2507~\citep{yang2025qwen3}. 
We construct the training corpus from 660K prompts from Infinity-Instruct-0625 dataset~\citep{li2025infinityinstruct} by generating responses with the corresponding target model. 
This ensures that the drafter training data distribution matches the target-model samples encountered at inference time.

We choose EAGLE-3~\citep{li2025eagle3} setup as the best-performing state-of-the-art draft training pipeline. 
All drafters are trained with $n=6$ speculative tokens with the weights shared across positions.
Draft backbone architecture is fixed for each target model, so the methods in the study differ only in the LM-head design. 
We employ standard KL-divergence loss as our training objective, unless stated otherwise.
More details on architectures, training hyperparameters and loss specifications are provided in Appendix~\ref{app:training_details}.

\subsection{Evaluation protocol}\label{subsec:eval_protocol}
We evaluate all methods across three benchmarks: \textbf{MT-Bench}~\citep{zheng2023mtbench}, \textbf{HumanEval}~\citep{chen2021humaneval}, and \textbf{GSM8K}~\citep{cobbe2021gsm8k}, covering such domains as instruction following, code generation and mathematical reasoning respectively. 
The evaluations are performed under both greedy (temperature $=0$) and stochastic decoding (temperature $=1$) with batch sizes $1$ and $64$, corresponding to latency-critical and high-throughput serving regimes.
We use production-like inference environment based on vLLM~0.17.1 with NVIDIA H200 GPUs.  

As our primary metric, we select generation throughput measured in tokens per second (TPS), which captures the end-to-end serving efficiency of each speculative decoding variant. 
To assess drafter quality independently of raw throughput, we additionally report average acceptance length $\tau$  as defined by~\eqref{eq:tau}. 
Each reported speedup value and $\tau$ is obtained by averaging over 5 identical runs with different random seeds.

\section{Evaluation Results}\label{sec:eval_results}

\begin{figure}[t]
  \centering
  \includegraphics[width=\linewidth]{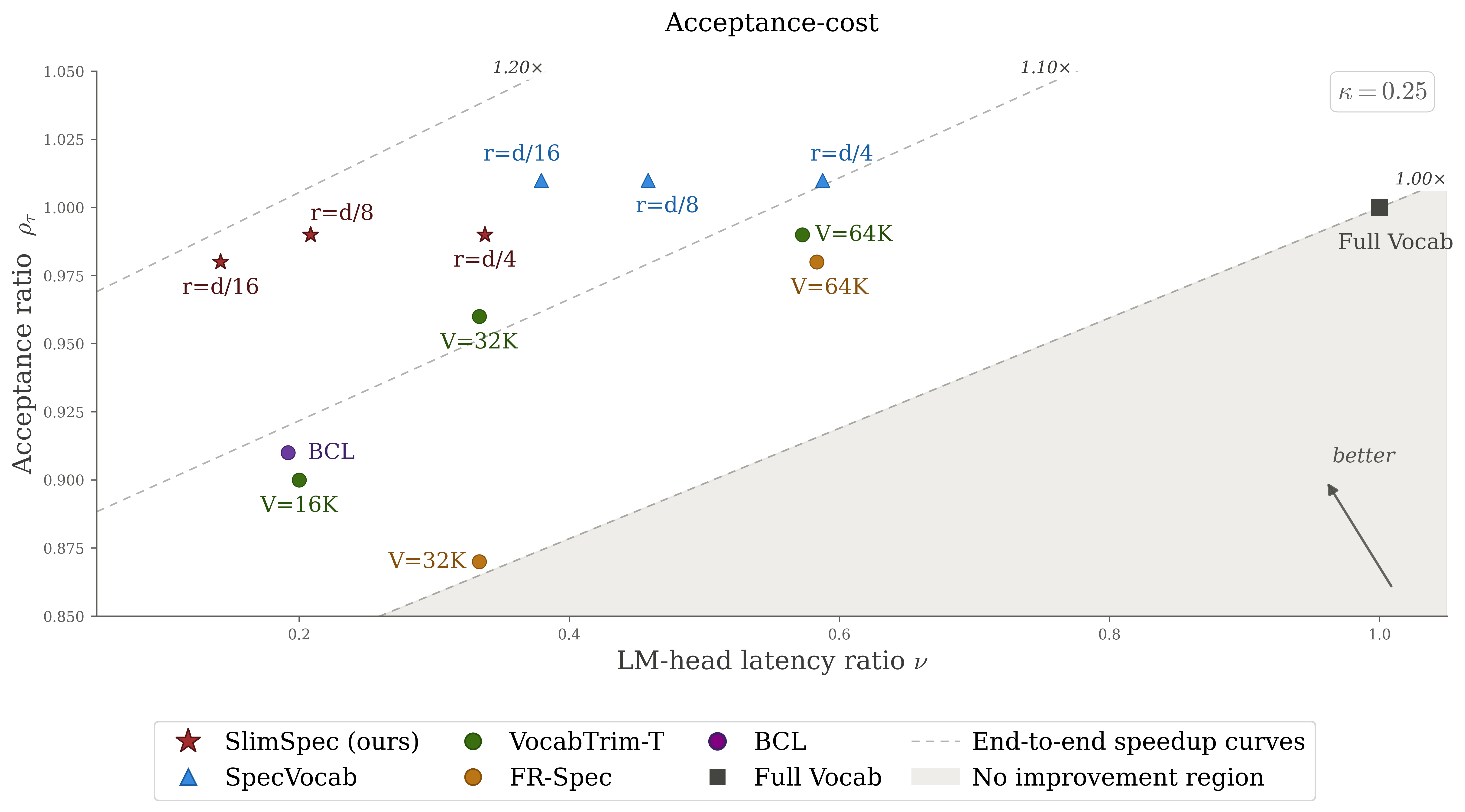}
  \caption{
  End-to-end speedup decomposition in the $(\nu, \rho_\tau)$ plane for Llama-3.1-8B with temperature $0$ at batch size $1$ ($\kappa = 0.25$).
  Dashed lines are theoretical speedup level curves derived from equation \ref{eq:speedup}. 
  The shaded region indicates no end-to-end improvement over the full-vocabulary baseline.
  SlimSpec (red stars) achieves the largest LM-head acceleration while keeping $\rho_\tau$ close to $1$. 
  }\label{fig:speedup_curves}
\end{figure}

As discussed in Section~\ref{subsec:speedup_improvement}, the efficiency of a draft-head design is governed by the trade-off between acceptance preservation $\rho_\tau$ and relative LM-head cost $\nu$.
In this section, we compare methods outlined in~\ref{subsec:methods} with Llama-3.1-8B as the target model and analyze their acceptance-cost trade-off in the $(\nu,\rho_\tau)$ plane.
Figure~\ref{fig:speedup_curves} plots the results for a representative subset of methods, with corresponding numerical values for temperature $0$ at batch size $1$ being reported in Appendix~\ref{app:acceptance_cost}.


Static vocabulary truncation (VocabTrim, FR-Spec, BCL) faces a clear cost-acceptance frontier: smaller vocabularies reduce $\nu$ but degrade $\rho_\tau$ proportionally.
FR-Spec is surpassed by VocabTrim because its frequency statistics, computed on a generic corpus, are less aligned with target-model generations than statistics collected on actual training data.
BCL selects too aggressive truncation strategy whose acceptance preservation is not compensated by the LM-head cost reduction.
Training-aware truncation (VocabTrim-T) demonstrates similar results to post-training truncation (VocabTrim) at any of the evaluated vocabulary sizes.
For clarity we plot only VocabTrim-T in Figure~\ref{fig:speedup_curves}.




Dynamic truncation with SpecVocab pushes the frontier upward, outperforming the static truncation baselines as expected.
With ranks $r\!=\!d/8$ and $r\!=\!d/16$ it preserves acceptance of the Full Vocab ($\rho_\tau\!\approx\!1$) while reducing LM-head latency approximately by $60\%$.

SlimSpec surpasses both these families.
With rank $r\!=\!d/8$ it reaches approximately $5 \times$ reduction in LM-head cost maintaining sufficiently high acceptance quality $\rho_\tau\!=\!0.99$. 
We adopt $r\!=\!d/8$ as the default SlimSpec configuration in the rest of the evaluation.

Finally, we compare SlimSpec against the strongest representative of each baseline family: Full Vocab, VocabTrim-T with $V_{\mathrm{tr}}\!=\!64\text{K}$, and SpecVocab with $r\!=\!d/8$.
Comparison summary is presented in Table~\ref{tab:main}, with detailed results being reported in Appendix~\ref{app:all_results}.
Averaged across benchmarks, it improves by more than $8.5\%$ over the static truncation baseline for Llama-3.1-8B with both batch sizes $1$ and $64$, and by $8.9\%$ over SpecVocab on GPT-OSS-20B at batch size $64$. 
For Qwen3-30B-A3B $T_{\mathrm{head}}$ accounts for a smaller fraction of the total latency, decreasing $\kappa$ significantly and shrinking the observed speedup to at most $1\text{-}2\%$ over the strongest baseline.
Overall, SlimSpec is never worse than the baselines in most setups, requiring no vocabulary curation or additional inference-time logic.

\begin{table}[t]
\centering
\caption{
    Comparison of SlimSpec with the main baselines for three target models on three benchmarks at temperature $0$.
    Speedup is measured relative to vanilla inference without a speculator.
    MT, HE, and GSM denote MT-Bench, HumanEval, and GSM8K, respectively.
    Avg denotes the arithmetic mean across these three benchmarks for each batch size.
}
\label{tab:main}
\setlength{\tabcolsep}{4pt}
\renewcommand{\arraystretch}{1.08}
\begin{tabular}{l CCC>{\columncolor{gray!12}}CCCC>{\columncolor{gray!12}}C}

\toprule
\multirow{2}{*}{Method}
& \multicolumn{4}{c}{Batch size $=1$}
& \multicolumn{4}{c}{Batch size $=64$} \\
\cmidrule(lr){2-5} \cmidrule(lr){6-9}
& MT & HE & GSM & Avg & MT & HE & GSM & Avg \\
\midrule
\multicolumn{9}{c}{\textbf{Llama-3.1-8B-Instruct}} \\
\midrule
Full Vocab & 2.23 & 2.78 & 2.42 & 2.48 & 1.35 & 1.43 & 1.41 & 1.40 \\
VocabTrim-T & 2.47 & 2.96 & 2.66 & 2.70 & 1.29 & 1.48 & 1.46 & 1.41 \\
SpecVocab & 2.59 & 3.17 & 2.82 & 2.86 & 1.36 & 1.50 & 1.52 & 1.46 \\
SlimSpec & {\bf 2.60} & {\bf 3.30} & {\bf 2.91} & {\bf 2.94} & {\bf 1.43} & {\bf 1.58} & {\bf 1.56} & {\bf 1.52} \\
\midrule
\multicolumn{9}{c}{\textbf{GPT-OSS-20B}} \\
\midrule
Full Vocab & 1.43 & 1.41 & 1.63 & 1.49 & 1.36 & 1.17 & 1.38 & 1.30 \\
VocabTrim-T & 1.72 & 1.67 & 1.95 & 1.78 & 1.50 & 1.27 & 1.52 & 1.43 \\
SpecVocab & 1.74 & 1.71 & 1.98 & 1.81 & 1.43 & 1.18 & 1.45 & 1.35 \\
SlimSpec & {\bf 1.76} & {\bf 1.76} & {\bf 2.11} & {\bf 1.88} & {\bf 1.52} & {\bf 1.34} & {\bf 1.56} & {\bf 1.47} \\
\midrule
\multicolumn{9}{c}{\textbf{Qwen3-30B-A3B-Instruct}} \\
\midrule
Full Vocab & 1.55 & 2.07 & 2.12 & 1.91 & 1.22 & 1.32 & 1.48 & 1.34 \\
VocabTrim-T & {\bf 1.66} & 2.15 & 2.23 & 2.01 & {\bf 1.31} & {\bf 1.36} & 1.53 & {\bf 1.40} \\
SpecVocab & {\bf 1.66} & 2.19 & 2.24 & 2.03 & 1.28 & 1.34 & 1.49 & 1.37 \\
SlimSpec & {\bf 1.66} & {\bf 2.22} & {\bf 2.27} & {\bf 2.05} & {\bf 1.31} & {\bf 1.36} & {\bf 1.54} & {\bf 1.40} \\
\bottomrule
\end{tabular}
\end{table}

\section{Conclusion}

In this work, we study different approaches to draft LM-head acceleration through the lens of the acceptance-cost trade-off rather than through the head latency alone.
Our performance model shows that a computationally cheaper head does not necessarily improve the end-to-end throughput.
This happens when the reduction in head cost is not offset by a loss in the average number of accepted draft tokens.
This perspective explains why vocabulary reduction methods require a careful design choice.
They make the final projection cheaper, but risk deteriorating acceptance quality or introducing new overheads, which limits the realized throughput gain.

We further present SlimSpec, an approach that follows a complementary direction, compressing the hidden representation predicted by the drafter backbone.
It preserves full-vocabulary support and moves primarily along the cost
axis, making the reduction in head latency more likely to translate into
end-to-end gains.
Our experiments confirm this intuition.
In the acceptance-cost plane, SlimSpec occupies the most favorable region among the evaluated draft-head designs, combining a substantially lower LM-head cost with acceptance close to the full-vocabulary baseline.
In our test settings, the rank-$d/8$ configuration gives the best overall trade-off, achieving approximately $4$-$5\times$ LM-head acceleration and the strongest throughput among the compared methods.

In our vision, future work would focus less on further reducing the head cost and more on improving acceptance at the same cost.
Therefore, acceptance-oriented training objectives are a natural candidate, since they could shift SlimSpec upward in the acceptance-cost plane without changing inference-time complexity.
Another direction could be position-wise adaptivity, e.g. by sharing the vocabulary-side projection while using position-specific compression projections, or by assigning larger ranks to positions where additional capacity has the highest impact.
Such extensions preserve the main advantage of SlimSpec while being driven by the acceptance-cost trade-off.

\section{Limitations}\label{sec:limitations}
The method proposed in this paper has a number of limitations, including:
(i) the rank $r$ is a manually chosen hyperparameter and we do not provide an automatic selection procedure;
(ii) all experiments are conducted with EAGLE-3 draft heads, so the conclusions drawn in the paper transfer to other drafter families (e.g. MEDUSA, Hydra) only by analogy;
(iii) measured speedups depend on the inference framework (vLLM 0.17.1) and hardware (NVIDIA H200) and may differ on other stacks; and
(iv) some of the dynamic-vocabulary methods, namely CORAL and DynaSpec, are not considered due to the lack of reference implementations, which narrows the coverage of our study.
We further discuss these limitations in detail in Appendix~\ref{app:limitations}.

\bibliographystyle{unsrtnat}
\bibliography{references}

\appendix
\clearpage
\section{Training Configurations}
\label{app:training_details}

All draft models are trained for 10 epochs with batch size 64 and learning rate \(4 \times 10^{-4}\). 
We use AdamW with \((\beta_1, \beta_2) = (0.9, 0.95)\), \(\epsilon = 10^{-8}\), and no weight decay. The learning rate is scheduled with a cosine decay after
\(100\) warmup steps, and gradients are clipped to a maximum norm of \(0.5\).

Each drafter follows the EAGLE-3-style architecture with \(n=6\) decoding heads; the models differ only in the design of the LM-head. 
All drafters except SpecVocab are trained with the forward KL objective:
\[
\mathrm{KL}(p\|q) = \sum_{v\in\mathcal{V}}p(v)\log\frac{p(v)}{q(v)},
\]
where $p$ and $q$ and target and draft model distributions respectively. 
For VocabTrim-T and BCL we re-normalize $p$ and take the sum over $\mathcal{V}_{\mathrm{tr}}$.
For SpecVocab, we follow the objective from the corresponding paper and add an auxiliary KL term with weight coefficient \(\lambda = 0.01\).

EAGLE-3 draft backbone consist of a single dense transformer layer that mirrors the target model’s architecture.
For MoE target models the intermediate dimension of the feed-forward network is chosen as
\[
d_{\text{ffn}} = \text{num-experts-per-block} \times d_{\text{expert}},
\]
where $\text{num-experts-per-block}$ is the number of experts activated per token and $d_{\text{expert}}$ is the intermediate dimension of each expert’s FFN. 

Training a single drafter requires 128 GPU-hours for Llama-3.1-8B-Instruct, 288 GPU-hours for GPT-OSS-20B, and 320 GPU-hours for Qwen3-30B-A3B-Instruct. 
All computations are performed on NVIDIA H200 GPUs.

\clearpage
\section{Limitations discussion}\label{app:limitations}

Our analysis and experiments focus on the regime where the draft LM head is a
non-negligible part of the speculative decoding pipeline.
The acceptance--cost model in Section~\ref{sec:performance_model} shows that LM-head acceleration translates into end-to-end speedup only when the saved head latency is large enough relative to the rest of the pipeline.
Consequently, the gains of SlimSpec are dependent on the deployment setup.
Serving stacks, GPU kernels, batch size, draft verification costs, sampling overhead and scheduler behavior can all change the effective value of $\kappa$ and therefore the realized throughput improvement.

Our empirical study is also limited to EAGLE-3 auxiliary drafters.
SlimSpec is a local replacement of the draft LM head and should be compatible with other auxiliary-drafter families such as MEDUSA or Hydra, but we do not evaluate those architectures directly.
Similarly, all main experiments use a fixed draft-backbone architecture and shared weights across draft positions.
We therefore do not characterize how the low-rank head interacts with alternative drafter backbones, dynamic draft trees, position-specific heads or standalone draft models.

The evaluation covers three target models and three benchmarks, but it does not exhaust the range of possible deployment scenarios.
We do not study really large target models (e.g. $>50$B parameters), long-context generation, multilingual or domain-specific workloads, tool-use settings, high sampling temperatures, alternative sampling policies, or other hardware and serving frameworks.
Since the acceptance--cost trade-off depends on both the target distribution and the inference stack, the numerical speedups reported here should be interpreted as measurements for the evaluated production-like setup rather than hardware-independent constants.

The rank $r$ is chosen manually, and we evaluate several simple fractions of the hidden dimension, finding that $r=d/8$ gives the best trade-off in our tested settings.
However, selecting the best rank depends on many factors, including the target model, vocabulary size, serving regime and benchmark distribution.
We do not provide an automatic rank-selection procedure or recommendations how to determine the optimal rank besides thorough hyperparameter tuning.

Finally, our comparison focuses on representative methods that we could implement faithfully under a shared training and inference pipeline.
We include static truncation methods and SpecVocab as a strong dynamic-vocabulary baseline, but we do not reproduce CORAL or DynaSpec because they require additional machinery, while no reference implementation was available to us.
More optimized kernels for dynamic vocabulary selection could change the relative overheads of these methods.

\clearpage
\section{Detailed acceptance--cost results}
\label{app:acceptance_cost}

\begin{table}[htbp]
\centering
\caption{
Acceptance--cost trade-off for all considered draft-head designs with Llama-3.1-8B at temperature $0$ and batch size $1$.
The table reports values averaged over three datasets: MT-Bench, HumanEval, and GSM8K.
Speedup is measured relative to the full-vocabulary baseline; $\rho_\tau$ denotes acceptance preservation, and $\nu$ denotes relative LM-head cost.
}
\label{tab:speedup_tau_nu_rel_full_vocab}
\setlength{\tabcolsep}{4pt}
\renewcommand{\arraystretch}{1.15}
\begin{tabular}{@{}llccc@{}}
\toprule
\multirow{2}{*}{\textbf{Method}}
& \multirow{2}{*}{\textbf{Configurations}}
& \multicolumn{3}{c}{\textbf{Batch size $=1$}} \\
\cmidrule(l){3-5}
&
& Speedup
& $\rho_\tau$
& $\nu$ \\
\midrule
\addlinespace[4pt]
Full Vocab & -- & $1.00$ & $1.00$ & $1.00$ \\
\addlinespace[4pt]

\multirow{3}{*}{FR-Spec} & $V_{\mathrm{tr}}=64$K & $1.06$ & $0.98$ & $0.58$ \\
 & $V_{\mathrm{tr}}=32$K & $1.01$ & $0.87$ & $0.33$ \\
 & $V_{\mathrm{tr}}=16$K & $0.93$ & $0.78$ & $0.20$ \\
\addlinespace[4pt]

\multirow{3}{*}{VocabTrim} & $V_{\mathrm{tr}}=64$K & $1.08$ & $0.99$ & $0.58$ \\
 & $V_{\mathrm{tr}}=32$K & $1.10$ & $0.96$ & $0.33$ \\
 & $V_{\mathrm{tr}}=16$K & $1.08$ & $0.90$ & $0.20$ \\
\addlinespace[4pt]

BCL & $V_{\mathrm{tr}}=15.8$K & $1.07$ & $0.91$ & $0.19$ \\
\addlinespace[4pt]

\multirow{3}{*}{VocabTrim-T} & $V_{\mathrm{tr}}=64$K & $1.09$ & $1.00$ & $0.58$ \\
 & $V_{\mathrm{tr}}=32$K & $1.09$ & $0.96$ & $0.33$ \\
 & $V_{\mathrm{tr}}=16$K & $1.07$ & $0.90$ & $0.20$ \\
\addlinespace[4pt]

\multirow{3}{*}{SpecVocab} & $r=d/4$ & $1.13$ & $1.01$ & $0.59$ \\
 & $r=d/8$ & $1.16$ & $1.01$ & $0.46$ \\
 & $r=d/16$ & $1.15$ & $1.01$ & $0.38$ \\
\addlinespace[4pt]

\multirow{3}{*}{SlimSpec} & $r=d/4$ & $1.16$ & $0.99$ & $0.34$ \\
 & $r=d/8$ & $1.19$ & $0.99$ & $0.21$ \\
 & $r=d/16$ & $1.18$ & $0.98$ & $0.14$ \\
\bottomrule
\end{tabular}
\end{table}


\clearpage
\clearpage
\section{Extended Results Tables}
\label{app:all_results}

We report vLLM speedup and average acceptance length measurements for all considered LM-head acceleration methods and their different configurations. The \textit{Config} column reports the corresponding hyperparameter for each method: it corresponds to the vocabulary size $V_\mathrm{tr}$ for methods with a static truncated vocabulary; , the intermediate router rank $r$ for SpecVocab; and the intermediate LM-head rank for SlimSpec.
\textit{Speedup} is measured relative to vanilla inference without a speculator, under the same benchmark, temperature, and batch size.

We provide 12 tables in total, covering all combinations of models, temperature settings, and batch sizes. For each benchmark and each configuration, we run the evaluation 5 times.  
We report Mean $\pm$ standard deviation over 5 these runs. 
\textbf{Bold} marks the best value among compressed methods, excluding \emph{Full Vocab}, which serves as the upper-bound reference.
\textit{Avg} denotes the arithmetic mean across these three benchmarks for each batch size.

\clearpage

\begin{table}[htbp]
\centering
\caption{Speedup for Llama3.1-8B-Instruct, temperature $=0$, batch size $=1$.}
\label{tab:llama3-1-8b-t0-bs1-speedup}

\setlength{\tabcolsep}{9pt}
\renewcommand{\arraystretch}{1.1}
\begin{tabular}{@{}ll ccc>{\columncolor{gray!12}}c@{}}
\toprule
\textbf{Method} & \textbf{Config} & \textbf{MT-Bench} & \textbf{Humaneval} & \textbf{GSM8K} & \textbf{Avg} \\
\midrule
Full Vocab & -- & 2.23 \Std{$\pm$ 0.02} & 2.78 \Std{$\pm$ 0.01} & 2.42 \Std{$\pm$ 0.01} & 2.48 \\
\addlinespace[2pt]

\multirow{3}{*}{FR-Spec} & $V_{\mathrm{tr}}=64$K & 2.35 \Std{$\pm$ 0.01} & 2.93 \Std{$\pm$ 0.02} & 2.59 \Std{$\pm$ 0.01} & 2.62 \\
 & $V_{\mathrm{tr}}=32$K & 2.27 \Std{$\pm$ 0.03} & 2.74 \Std{$\pm$ 0.02} & 2.47 \Std{$\pm$ 0.01} & 2.50 \\
 & $V_{\mathrm{tr}}=16$K & 2.09 \Std{$\pm$ 0.01} & 2.47 \Std{$\pm$ 0.01} & 2.34 \Std{$\pm$ 0.01} & 2.30 \\
\addlinespace[2pt]

\multirow{3}{*}{VocabTrim} & $V_{\mathrm{tr}}=64$K & 2.42 \Std{$\pm$ 0.02} & 3.01 \Std{$\pm$ 0.02} & 2.63 \Std{$\pm$ 0.02} & 2.69 \\
 & $V_{\mathrm{tr}}=32$K & 2.50 \Std{$\pm$ 0.02} & 2.98 \Std{$\pm$ 0.02} & 2.66 \Std{$\pm$ 0.03} & 2.71 \\
 & $V_{\mathrm{tr}}=16$K & 2.44 \Std{$\pm$ 0.01} & 2.91 \Std{$\pm$ 0.02} & 2.63 \Std{$\pm$ 0.02} & 2.66 \\
\addlinespace[2pt]

BCL & $V_{\mathrm{tr}}=15.8$K & 2.45 \Std{$\pm$ 0.01} & 2.89 \Std{$\pm$ 0.02} & 2.63 \Std{$\pm$ 0.00} & 2.65 \\
\addlinespace[2pt]

\multirow{3}{*}{VocabTrim-T} & $V_{\mathrm{tr}}=64$K & 2.47 \Std{$\pm$ 0.03} & 2.96 \Std{$\pm$ 0.02} & 2.66 \Std{$\pm$ 0.01} & 2.70 \\
 & $V_{\mathrm{tr}}=32$K & 2.50 \Std{$\pm$ 0.02} & 2.96 \Std{$\pm$ 0.02} & 2.65 \Std{$\pm$ 0.01} & 2.70 \\
 & $V_{\mathrm{tr}}=16$K & 2.43 \Std{$\pm$ 0.05} & 2.88 \Std{$\pm$ 0.03} & 2.64 \Std{$\pm$ 0.02} & 2.65 \\
\addlinespace[2pt]

\multirow{3}{*}{SpecVocab} & $r=d/4$ & 2.50 \Std{$\pm$ 0.03} & 3.13 \Std{$\pm$ 0.02} & 2.74 \Std{$\pm$ 0.01} & 2.79 \\
 & $r=d/8$ & 2.59 \Std{$\pm$ 0.02} & 3.17 \Std{$\pm$ 0.02} & 2.82 \Std{$\pm$ 0.01} & 2.86 \\
 & $r=d/16$ & 2.59 \Std{$\pm$ 0.03} & 3.16 \Std{$\pm$ 0.03} & 2.80 \Std{$\pm$ 0.03} & 2.85 \\
\midrule
\multirow{3}{*}{SlimSpec} & $r=d/4$ & 2.58 \Std{$\pm$ 0.02} & 3.22 \Std{$\pm$ 0.02} & 2.81 \Std{$\pm$ 0.02} & 2.87 \\
 & $r=d/8$ & 2.60 \Std{$\pm$ 0.03} & {\bf 3.30} \Std{$\pm$ 0.03} & {\bf 2.91} \Std{$\pm$ 0.02} & {\bf 2.94} \\
 & $r=d/16$ & {\bf 2.61} \Std{$\pm$ 0.01} & 3.29 \Std{$\pm$ 0.03} & 2.90 \Std{$\pm$ 0.00} & 2.93 \\
\bottomrule
\end{tabular}
\end{table}

\begin{table}[htbp]
\centering
\caption{Average acceptance length $\tau$ for Llama3.1-8B-Instruct, temperature $=0$, batch size $=1$.}
\label{tab:llama3-1-8b-t0-bs1-tau}

\setlength{\tabcolsep}{9pt}
\renewcommand{\arraystretch}{1.1}
\begin{tabular}{@{}ll ccc>{\columncolor{gray!12}}c@{}}
\toprule
\textbf{Method} & \textbf{Config} & \textbf{MT-Bench} & \textbf{Humaneval} & \textbf{GSM8K} & \textbf{Avg} \\
\midrule
Full Vocab & -- & 3.93 \Std{$\pm$ 0.00} & 4.93 \Std{$\pm$ 0.02} & 4.40 \Std{$\pm$ 0.00} & 4.42 \\
\addlinespace[2pt]

\multirow{3}{*}{FR-Spec} & $V_{\mathrm{tr}}=64$K & 3.81 \Std{$\pm$ 0.00} & 4.80 \Std{$\pm$ 0.02} & 4.33 \Std{$\pm$ 0.00} & 4.31 \\
 & $V_{\mathrm{tr}}=32$K & 3.46 \Std{$\pm$ 0.01} & 4.23 \Std{$\pm$ 0.01} & 3.89 \Std{$\pm$ 0.00} & 3.86 \\
 & $V_{\mathrm{tr}}=16$K & 3.08 \Std{$\pm$ 0.01} & 3.71 \Std{$\pm$ 0.00} & 3.59 \Std{$\pm$ 0.00} & 3.46 \\
\addlinespace[2pt]

\multirow{3}{*}{VocabTrim} & $V_{\mathrm{tr}}=64$K & 3.92 \Std{$\pm$ 0.00} & 4.89 \Std{$\pm$ 0.03} & 4.38 \Std{$\pm$ 0.00} & 4.40 \\
 & $V_{\mathrm{tr}}=32$K & 3.84 \Std{$\pm$ 0.01} & 4.64 \Std{$\pm$ 0.02} & 4.26 \Std{$\pm$ 0.00} & 4.25 \\
 & $V_{\mathrm{tr}}=16$K & 3.60 \Std{$\pm$ 0.01} & 4.33 \Std{$\pm$ 0.02} & 4.04 \Std{$\pm$ 0.00} & 3.99 \\
\addlinespace[2pt]

BCL & $V_{\mathrm{tr}}=15.8$K & 3.64 \Std{$\pm$ 0.01} & 4.34 \Std{$\pm$ 0.01} & 4.08 \Std{$\pm$ 0.01} & 4.02 \\
\addlinespace[2pt]

\multirow{3}{*}{VocabTrim-T} & $V_{\mathrm{tr}}=64$K & 3.95 \Std{$\pm$ 0.01} & 4.88 \Std{$\pm$ 0.01} & 4.37 \Std{$\pm$ 0.01} & 4.40 \\
 & $V_{\mathrm{tr}}=32$K & 3.86 \Std{$\pm$ 0.01} & 4.61 \Std{$\pm$ 0.01} & 4.24 \Std{$\pm$ 0.01} & 4.24 \\
 & $V_{\mathrm{tr}}=16$K & 3.63 \Std{$\pm$ 0.02} & 4.29 \Std{$\pm$ 0.01} & 4.04 \Std{$\pm$ 0.00} & 3.99 \\
\addlinespace[2pt]

\multirow{3}{*}{SpecVocab} & $r=d/4$ & 3.94 \Std{$\pm$ 0.01} & 4.95 \Std{$\pm$ 0.01} & 4.45 \Std{$\pm$ 0.00} & 4.45 \\
 & $r=d/8$ & 3.96 \Std{$\pm$ 0.02} & 4.93 \Std{$\pm$ 0.02} & 4.47 \Std{$\pm$ 0.01} & 4.45 \\
 & $r=d/16$ & 3.99 \Std{$\pm$ 0.02} & 4.93 \Std{$\pm$ 0.03} & 4.45 \Std{$\pm$ 0.00} & 4.46 \\
\midrule
\multirow{3}{*}{SlimSpec} & $r=d/4$ & 3.90 \Std{$\pm$ 0.03} & 4.87 \Std{$\pm$ 0.03} & 4.39 \Std{$\pm$ 0.00} & 4.39 \\
 & $r=d/8$ & 3.84 \Std{$\pm$ 0.02} & 4.89 \Std{$\pm$ 0.01} & 4.39 \Std{$\pm$ 0.00} & 4.37 \\
 & $r=d/16$ & 3.79 \Std{$\pm$ 0.01} & 4.83 \Std{$\pm$ 0.03} & 4.37 \Std{$\pm$ 0.01} & 4.33 \\
\bottomrule
\end{tabular}
\end{table}

\clearpage
\begin{table}[htbp]
\centering
\caption{Speedup for Llama3.1-8B-Instruct, temperature $=0$, batch size $=64$.}
\label{tab:llama3-1-8b-t0-bs64-speedup}

\setlength{\tabcolsep}{9pt}
\renewcommand{\arraystretch}{1.1}
\begin{tabular}{@{}ll ccc>{\columncolor{gray!12}}c@{}}
\toprule
\textbf{Method} & \textbf{Config} & \textbf{MT-Bench} & \textbf{Humaneval} & \textbf{GSM8K} & \textbf{Avg} \\
\midrule
Full Vocab & -- & 1.35 \Std{$\pm$ 0.01} & 1.43 \Std{$\pm$ 0.01} & 1.41 \Std{$\pm$ 0.02} & 1.39 \\
\addlinespace[2pt]

\multirow{3}{*}{FR-Spec} & $V_{\mathrm{tr}}=64$K & 1.35 \Std{$\pm$ 0.02} & 1.49 \Std{$\pm$ 0.03} & 1.45 \Std{$\pm$ 0.02} & 1.43 \\
 & $V_{\mathrm{tr}}=32$K & 1.33 \Std{$\pm$ 0.01} & 1.45 \Std{$\pm$ 0.01} & 1.41 \Std{$\pm$ 0.02} & 1.40 \\
 & $V_{\mathrm{tr}}=16$K & 1.19 \Std{$\pm$ 0.01} & 1.32 \Std{$\pm$ 0.02} & 1.35 \Std{$\pm$ 0.02} & 1.29 \\
\addlinespace[2pt]

\multirow{3}{*}{VocabTrim} & $V_{\mathrm{tr}}=64$K & 1.37 \Std{$\pm$ 0.02} & 1.51 \Std{$\pm$ 0.02} & 1.49 \Std{$\pm$ 0.03} & 1.46 \\
 & $V_{\mathrm{tr}}=32$K & 1.39 \Std{$\pm$ 0.04} & 1.44 \Std{$\pm$ 0.03} & 1.49 \Std{$\pm$ 0.02} & 1.44 \\
 & $V_{\mathrm{tr}}=16$K & 1.35 \Std{$\pm$ 0.02} & 1.46 \Std{$\pm$ 0.04} & 1.48 \Std{$\pm$ 0.02} & 1.43 \\
\addlinespace[2pt]

BCL & $V_{\mathrm{tr}}=15.8$K & 1.37 \Std{$\pm$ 0.02} & 1.48 \Std{$\pm$ 0.02} & 1.47 \Std{$\pm$ 0.01} & 1.44 \\
\addlinespace[2pt]

\multirow{3}{*}{VocabTrim-T} & $V_{\mathrm{tr}}=64$K & 1.29 \Std{$\pm$ 0.03} & 1.48 \Std{$\pm$ 0.03} & 1.46 \Std{$\pm$ 0.01} & 1.41 \\
 & $V_{\mathrm{tr}}=32$K & 1.29 \Std{$\pm$ 0.03} & 1.48 \Std{$\pm$ 0.03} & 1.45 \Std{$\pm$ 0.02} & 1.41 \\
 & $V_{\mathrm{tr}}=16$K & 1.37 \Std{$\pm$ 0.03} & 1.49 \Std{$\pm$ 0.02} & 1.46 \Std{$\pm$ 0.03} & 1.44 \\
\addlinespace[2pt]

\multirow{3}{*}{SpecVocab} & $r=d/4$ & 1.36 \Std{$\pm$ 0.02} & 1.39 \Std{$\pm$ 0.20} & 1.44 \Std{$\pm$ 0.05} & 1.40 \\
 & $r=d/8$ & 1.36 \Std{$\pm$ 0.04} & 1.50 \Std{$\pm$ 0.04} & 1.52 \Std{$\pm$ 0.04} & 1.46 \\
 & $r=d/16$ & 1.36 \Std{$\pm$ 0.03} & 1.50 \Std{$\pm$ 0.03} & 1.48 \Std{$\pm$ 0.09} & 1.45 \\
\midrule
\multirow{3}{*}{SlimSpec} & $r=d/4$ & 1.41 \Std{$\pm$ 0.05} & 1.56 \Std{$\pm$ 0.04} & 1.55 \Std{$\pm$ 0.04} & 1.51 \\
 & $r=d/8$ & {\bf 1.43} \Std{$\pm$ 0.04} & {\bf 1.58} \Std{$\pm$ 0.05} & {\bf 1.56} \Std{$\pm$ 0.03} & {\bf 1.53} \\
 & $r=d/16$ & 1.41 \Std{$\pm$ 0.01} & 1.56 \Std{$\pm$ 0.05} & 1.50 \Std{$\pm$ 0.03} & 1.49 \\
\bottomrule
\end{tabular}
\end{table}

\begin{table}[htbp]
\centering
\caption{Average acceptance length $\tau$ for Llama3.1-8B-Instruct, temperature $=0$, batch size $=64$.}
\label{tab:llama3-1-8b-t0-bs64-tau}

\setlength{\tabcolsep}{9pt}
\renewcommand{\arraystretch}{1.1}
\begin{tabular}{@{}ll ccc>{\columncolor{gray!12}}c@{}}
\toprule
\textbf{Method} & \textbf{Config} & \textbf{MT-Bench} & \textbf{Humaneval} & \textbf{GSM8K} & \textbf{Avg} \\
\midrule
Full Vocab & -- & 3.91 \Std{$\pm$ 0.01} & 4.90 \Std{$\pm$ 0.02} & 4.38 \Std{$\pm$ 0.03} & 4.40 \\
\addlinespace[2pt]

\multirow{3}{*}{FR-Spec} & $V_{\mathrm{tr}}=64$K & 3.76 \Std{$\pm$ 0.02} & 4.77 \Std{$\pm$ 0.02} & 4.31 \Std{$\pm$ 0.00} & 4.28 \\
 & $V_{\mathrm{tr}}=32$K & 3.43 \Std{$\pm$ 0.02} & 4.22 \Std{$\pm$ 0.01} & 3.85 \Std{$\pm$ 0.02} & 3.83 \\
 & $V_{\mathrm{tr}}=16$K & 3.06 \Std{$\pm$ 0.02} & 3.71 \Std{$\pm$ 0.01} & 3.57 \Std{$\pm$ 0.02} & 3.45 \\
\addlinespace[2pt]

\multirow{3}{*}{VocabTrim} & $V_{\mathrm{tr}}=64$K & 3.88 \Std{$\pm$ 0.04} & 4.89 \Std{$\pm$ 0.02} & 4.35 \Std{$\pm$ 0.02} & 4.37 \\
 & $V_{\mathrm{tr}}=32$K & 3.80 \Std{$\pm$ 0.02} & 4.63 \Std{$\pm$ 0.03} & 4.22 \Std{$\pm$ 0.02} & 4.22 \\
 & $V_{\mathrm{tr}}=16$K & 3.58 \Std{$\pm$ 0.01} & 4.33 \Std{$\pm$ 0.02} & 4.02 \Std{$\pm$ 0.00} & 3.98 \\
\addlinespace[2pt]

BCL & $V_{\mathrm{tr}}=15.8$K & 3.60 \Std{$\pm$ 0.01} & 4.32 \Std{$\pm$ 0.01} & 4.06 \Std{$\pm$ 0.01} & 3.99 \\
\addlinespace[2pt]

\multirow{3}{*}{VocabTrim-T} & $V_{\mathrm{tr}}=64$K & 3.93 \Std{$\pm$ 0.02} & 4.88 \Std{$\pm$ 0.03} & 4.37 \Std{$\pm$ 0.02} & 4.39 \\
 & $V_{\mathrm{tr}}=32$K & 3.82 \Std{$\pm$ 0.02} & 4.60 \Std{$\pm$ 0.01} & 4.24 \Std{$\pm$ 0.01} & 4.22 \\
 & $V_{\mathrm{tr}}=16$K & 3.59 \Std{$\pm$ 0.02} & 4.29 \Std{$\pm$ 0.02} & 4.03 \Std{$\pm$ 0.02} & 3.97 \\
\addlinespace[2pt]

\multirow{3}{*}{SpecVocab} & $r=d/4$ & 3.94 \Std{$\pm$ 0.02} & 4.93 \Std{$\pm$ 0.01} & 4.43 \Std{$\pm$ 0.02} & 4.43 \\
 & $r=d/8$ & 3.92 \Std{$\pm$ 0.02} & 4.93 \Std{$\pm$ 0.02} & 4.45 \Std{$\pm$ 0.01} & 4.43 \\
 & $r=d/16$ & 3.96 \Std{$\pm$ 0.01} & 4.93 \Std{$\pm$ 0.01} & 4.43 \Std{$\pm$ 0.02} & 4.44 \\
\midrule
\multirow{3}{*}{SlimSpec} & $r=d/4$ & 3.87 \Std{$\pm$ 0.01} & 4.89 \Std{$\pm$ 0.02} & 4.39 \Std{$\pm$ 0.02} & 4.38 \\
 & $r=d/8$ & 3.81 \Std{$\pm$ 0.02} & 4.89 \Std{$\pm$ 0.02} & 4.39 \Std{$\pm$ 0.02} & 4.36 \\
 & $r=d/16$ & 3.78 \Std{$\pm$ 0.02} & 4.83 \Std{$\pm$ 0.01} & 4.36 \Std{$\pm$ 0.01} & 4.32 \\
\bottomrule
\end{tabular}
\end{table}

\clearpage
\begin{table}[htbp]
\centering
\caption{Speedup for Llama3.1-8B-Instruct, temperature $=1$, batch size $=1$.}
\label{tab:llama3-1-8b-t1-bs1-speedup}

\setlength{\tabcolsep}{9pt}
\renewcommand{\arraystretch}{1.1}
\begin{tabular}{@{}ll ccc>{\columncolor{gray!12}}c@{}}
\toprule
\textbf{Method} & \textbf{Config} & \textbf{MT-Bench} & \textbf{Humaneval} & \textbf{GSM8K} & \textbf{Avg} \\
\midrule
Full Vocab & -- & 1.83 \Std{$\pm$ 0.02} & 2.33 \Std{$\pm$ 0.03} & 2.02 \Std{$\pm$ 0.03} & 2.06 \\
\addlinespace[2pt]

\multirow{3}{*}{FR-Spec} & $V_{\mathrm{tr}}=64$K & 1.88 \Std{$\pm$ 0.03} & 2.34 \Std{$\pm$ 0.03} & 2.00 \Std{$\pm$ 0.02} & 2.07 \\
 & $V_{\mathrm{tr}}=32$K & 1.75 \Std{$\pm$ 0.03} & 2.09 \Std{$\pm$ 0.05} & 1.85 \Std{$\pm$ 0.01} & 1.90 \\
 & $V_{\mathrm{tr}}=16$K & 1.56 \Std{$\pm$ 0.01} & 1.84 \Std{$\pm$ 0.01} & 1.70 \Std{$\pm$ 0.04} & 1.70 \\
\addlinespace[2pt]

\multirow{3}{*}{VocabTrim} & $V_{\mathrm{tr}}=64$K & 1.91 \Std{$\pm$ 0.03} & 2.38 \Std{$\pm$ 0.05} & 2.05 \Std{$\pm$ 0.05} & 2.11 \\
 & $V_{\mathrm{tr}}=32$K & 1.83 \Std{$\pm$ 0.05} & 2.29 \Std{$\pm$ 0.02} & 2.01 \Std{$\pm$ 0.02} & 2.04 \\
 & $V_{\mathrm{tr}}=16$K & 1.78 \Std{$\pm$ 0.02} & 2.14 \Std{$\pm$ 0.03} & 1.94 \Std{$\pm$ 0.02} & 1.95 \\
\addlinespace[2pt]

BCL & $V_{\mathrm{tr}}=15.8$K & 1.76 \Std{$\pm$ 0.04} & 2.18 \Std{$\pm$ 0.02} & 2.02 \Std{$\pm$ 0.04} & 1.99 \\
\addlinespace[2pt]

\multirow{3}{*}{VocabTrim-T} & $V_{\mathrm{tr}}=64$K & 1.93 \Std{$\pm$ 0.04} & 2.39 \Std{$\pm$ 0.04} & 2.11 \Std{$\pm$ 0.01} & 2.14 \\
 & $V_{\mathrm{tr}}=32$K & 1.92 \Std{$\pm$ 0.03} & 2.29 \Std{$\pm$ 0.03} & 2.01 \Std{$\pm$ 0.02} & 2.08 \\
 & $V_{\mathrm{tr}}=16$K & 1.80 \Std{$\pm$ 0.05} & 2.10 \Std{$\pm$ 0.05} & 1.94 \Std{$\pm$ 0.03} & 1.95 \\
\addlinespace[2pt]

\multirow{3}{*}{SpecVocab} & $r=d/4$ & 2.10 \Std{$\pm$ 0.03} & 2.62 \Std{$\pm$ 0.03} & 2.24 \Std{$\pm$ 0.03} & 2.32 \\
 & $r=d/8$ & 2.11 \Std{$\pm$ 0.05} & 2.66 \Std{$\pm$ 0.05} & 2.32 \Std{$\pm$ 0.02} & 2.36 \\
 & $r=d/16$ & {\bf 2.17} \Std{$\pm$ 0.04} & 2.67 \Std{$\pm$ 0.03} & 2.30 \Std{$\pm$ 0.02} & 2.38 \\
\midrule
\multirow{3}{*}{SlimSpec} & $r=d/4$ & 2.16 \Std{$\pm$ 0.02} & 2.68 \Std{$\pm$ 0.06} & {\bf 2.35} \Std{$\pm$ 0.01} & {\bf 2.40} \\
 & $r=d/8$ & 2.15 \Std{$\pm$ 0.02} & {\bf 2.71} \Std{$\pm$ 0.03} & 2.31 \Std{$\pm$ 0.06} & 2.39 \\
 & $r=d/16$ & 2.13 \Std{$\pm$ 0.02} & 2.70 \Std{$\pm$ 0.03} & 2.30 \Std{$\pm$ 0.06} & 2.37 \\
\bottomrule
\end{tabular}
\end{table}

\begin{table}[htbp]
\centering
\caption{Average acceptance length $\tau$ for Llama3.1-8B-Instruct, temperature $=1$, batch size $=1$.}
\label{tab:llama3-1-8b-t1-bs1-tau}

\setlength{\tabcolsep}{9pt}
\renewcommand{\arraystretch}{1.1}
\begin{tabular}{@{}ll ccc>{\columncolor{gray!12}}c@{}}
\toprule
\textbf{Method} & \textbf{Config} & \textbf{MT-Bench} & \textbf{Humaneval} & \textbf{GSM8K} & \textbf{Avg} \\
\midrule
Full Vocab & -- & 3.49 \Std{$\pm$ 0.03} & 4.46 \Std{$\pm$ 0.04} & 3.89 \Std{$\pm$ 0.03} & 3.95 \\
\addlinespace[2pt]

\multirow{3}{*}{FR-Spec} & $V_{\mathrm{tr}}=64$K & 3.41 \Std{$\pm$ 0.01} & 4.33 \Std{$\pm$ 0.04} & 3.86 \Std{$\pm$ 0.04} & 3.87 \\
 & $V_{\mathrm{tr}}=32$K & 3.17 \Std{$\pm$ 0.03} & 3.85 \Std{$\pm$ 0.04} & 3.50 \Std{$\pm$ 0.03} & 3.51 \\
 & $V_{\mathrm{tr}}=16$K & 2.87 \Std{$\pm$ 0.01} & 3.43 \Std{$\pm$ 0.02} & 3.25 \Std{$\pm$ 0.02} & 3.18 \\
\addlinespace[2pt]

\multirow{3}{*}{VocabTrim} & $V_{\mathrm{tr}}=64$K & 3.49 \Std{$\pm$ 0.02} & 4.39 \Std{$\pm$ 0.04} & 3.88 \Std{$\pm$ 0.04} & 3.92 \\
 & $V_{\mathrm{tr}}=32$K & 3.42 \Std{$\pm$ 0.03} & 4.24 \Std{$\pm$ 0.01} & 3.78 \Std{$\pm$ 0.05} & 3.81 \\
 & $V_{\mathrm{tr}}=16$K & 3.28 \Std{$\pm$ 0.02} & 3.95 \Std{$\pm$ 0.03} & 3.64 \Std{$\pm$ 0.02} & 3.62 \\
\addlinespace[2pt]

BCL & $V_{\mathrm{tr}}=15.8$K & 3.28 \Std{$\pm$ 0.03} & 3.91 \Std{$\pm$ 0.05} & 3.67 \Std{$\pm$ 0.03} & 3.62 \\
\addlinespace[2pt]

\multirow{3}{*}{VocabTrim-T} & $V_{\mathrm{tr}}=64$K & 3.49 \Std{$\pm$ 0.03} & 4.39 \Std{$\pm$ 0.02} & 3.90 \Std{$\pm$ 0.02} & 3.93 \\
 & $V_{\mathrm{tr}}=32$K & 3.40 \Std{$\pm$ 0.05} & 4.20 \Std{$\pm$ 0.02} & 3.74 \Std{$\pm$ 0.03} & 3.78 \\
 & $V_{\mathrm{tr}}=16$K & 3.26 \Std{$\pm$ 0.03} & 3.92 \Std{$\pm$ 0.04} & 3.63 \Std{$\pm$ 0.02} & 3.60 \\
\addlinespace[2pt]

\multirow{3}{*}{SpecVocab} & $r=d/4$ & 3.51 \Std{$\pm$ 0.05} & 4.45 \Std{$\pm$ 0.03} & 3.93 \Std{$\pm$ 0.03} & 3.96 \\
 & $r=d/8$ & 3.50 \Std{$\pm$ 0.05} & 4.45 \Std{$\pm$ 0.04} & 3.93 \Std{$\pm$ 0.04} & 3.96 \\
 & $r=d/16$ & 3.53 \Std{$\pm$ 0.04} & 4.43 \Std{$\pm$ 0.05} & 3.92 \Std{$\pm$ 0.03} & 3.96 \\
\midrule
\multirow{3}{*}{SlimSpec} & $r=d/4$ & 3.46 \Std{$\pm$ 0.04} & 4.41 \Std{$\pm$ 0.06} & 3.89 \Std{$\pm$ 0.01} & 3.92 \\
 & $r=d/8$ & 3.42 \Std{$\pm$ 0.03} & 4.42 \Std{$\pm$ 0.05} & 3.84 \Std{$\pm$ 0.05} & 3.89 \\
 & $r=d/16$ & 3.34 \Std{$\pm$ 0.02} & 4.34 \Std{$\pm$ 0.05} & 3.82 \Std{$\pm$ 0.03} & 3.83 \\
\bottomrule
\end{tabular}
\end{table}

\clearpage
\begin{table}[htbp]
\centering
\caption{Speedup for Llama3.1-8B-Instruct, temperature $=1$, batch size $=64$.}
\label{tab:llama3-1-8b-t1-bs64-speedup}

\setlength{\tabcolsep}{9pt}
\renewcommand{\arraystretch}{1.1}
\begin{tabular}{@{}ll ccc>{\columncolor{gray!12}}c@{}}
\toprule
\textbf{Method} & \textbf{Config} & \textbf{MT-Bench} & \textbf{Humaneval} & \textbf{GSM8K} & \textbf{Avg} \\
\midrule
Full Vocab & -- & 1.19 \Std{$\pm$ 0.03} & 1.28 \Std{$\pm$ 0.05} & 1.28 \Std{$\pm$ 0.03} & 1.25 \\
\addlinespace[2pt]

\multirow{3}{*}{FR-Spec} & $V_{\mathrm{tr}}=64$K & 1.21 \Std{$\pm$ 0.00} & 1.32 \Std{$\pm$ 0.03} & 1.32 \Std{$\pm$ 0.01} & 1.28 \\
 & $V_{\mathrm{tr}}=32$K & 1.16 \Std{$\pm$ 0.02} & 1.30 \Std{$\pm$ 0.03} & 1.26 \Std{$\pm$ 0.03} & 1.24 \\
 & $V_{\mathrm{tr}}=16$K & 1.08 \Std{$\pm$ 0.03} & 1.15 \Std{$\pm$ 0.04} & 1.21 \Std{$\pm$ 0.02} & 1.15 \\
\addlinespace[2pt]

\multirow{3}{*}{VocabTrim} & $V_{\mathrm{tr}}=64$K & 1.22 \Std{$\pm$ 0.03} & 1.36 \Std{$\pm$ 0.04} & 1.31 \Std{$\pm$ 0.03} & 1.30 \\
 & $V_{\mathrm{tr}}=32$K & 1.23 \Std{$\pm$ 0.05} & 1.32 \Std{$\pm$ 0.01} & 1.28 \Std{$\pm$ 0.04} & 1.27 \\
 & $V_{\mathrm{tr}}=16$K & 1.18 \Std{$\pm$ 0.02} & 1.31 \Std{$\pm$ 0.04} & 1.31 \Std{$\pm$ 0.04} & 1.27 \\
\addlinespace[2pt]

BCL & $V_{\mathrm{tr}}=15.8$K & 1.21 \Std{$\pm$ 0.05} & 1.29 \Std{$\pm$ 0.03} & 1.33 \Std{$\pm$ 0.04} & 1.28 \\
\addlinespace[2pt]

\multirow{3}{*}{VocabTrim-T} & $V_{\mathrm{tr}}=64$K & 1.23 \Std{$\pm$ 0.03} & 1.35 \Std{$\pm$ 0.04} & 1.30 \Std{$\pm$ 0.03} & 1.29 \\
 & $V_{\mathrm{tr}}=32$K & 1.19 \Std{$\pm$ 0.04} & 1.33 \Std{$\pm$ 0.02} & 1.30 \Std{$\pm$ 0.02} & 1.27 \\
 & $V_{\mathrm{tr}}=16$K & 1.16 \Std{$\pm$ 0.03} & 1.30 \Std{$\pm$ 0.03} & 1.30 \Std{$\pm$ 0.02} & 1.25 \\
\addlinespace[2pt]

\multirow{3}{*}{SpecVocab} & $r=d/4$ & 1.22 \Std{$\pm$ 0.05} & 1.24 \Std{$\pm$ 0.20} & 1.29 \Std{$\pm$ 0.06} & 1.25 \\
 & $r=d/8$ & 1.17 \Std{$\pm$ 0.04} & 1.30 \Std{$\pm$ 0.05} & 1.32 \Std{$\pm$ 0.05} & 1.26 \\
 & $r=d/16$ & 1.17 \Std{$\pm$ 0.04} & 1.19 \Std{$\pm$ 0.27} & 1.34 \Std{$\pm$ 0.03} & 1.24 \\
\midrule
\multirow{3}{*}{SlimSpec} & $r=d/4$ & 1.24 \Std{$\pm$ 0.05} & {\bf 1.42} \Std{$\pm$ 0.05} & 1.36 \Std{$\pm$ 0.06} & 1.34 \\
 & $r=d/8$ & 1.23 \Std{$\pm$ 0.06} & {\bf 1.42} \Std{$\pm$ 0.04} & {\bf 1.38} \Std{$\pm$ 0.05} & {\bf 1.35} \\
 & $r=d/16$ & {\bf 1.27} \Std{$\pm$ 0.03} & 1.36 \Std{$\pm$ 0.04} & 1.36 \Std{$\pm$ 0.05} & 1.33 \\
\bottomrule
\end{tabular}
\end{table}

\begin{table}[htbp]
\centering
\caption{Average acceptance length $\tau$ for Llama3.1-8B-Instruct, temperature $=1$, batch size $=64$.}
\label{tab:llama3-1-8b-t1-bs64-tau}

\setlength{\tabcolsep}{9pt}
\renewcommand{\arraystretch}{1.1}
\begin{tabular}{@{}ll ccc>{\columncolor{gray!12}}c@{}}
\toprule
\textbf{Method} & \textbf{Config} & \textbf{MT-Bench} & \textbf{Humaneval} & \textbf{GSM8K} & \textbf{Avg} \\
\midrule
Full Vocab & -- & 3.47 \Std{$\pm$ 0.03} & 4.43 \Std{$\pm$ 0.05} & 3.86 \Std{$\pm$ 0.03} & 3.92 \\
\addlinespace[2pt]

\multirow{3}{*}{FR-Spec} & $V_{\mathrm{tr}}=64$K & 3.38 \Std{$\pm$ 0.04} & 4.29 \Std{$\pm$ 0.04} & 3.80 \Std{$\pm$ 0.04} & 3.82 \\
 & $V_{\mathrm{tr}}=32$K & 3.13 \Std{$\pm$ 0.04} & 3.89 \Std{$\pm$ 0.04} & 3.45 \Std{$\pm$ 0.03} & 3.49 \\
 & $V_{\mathrm{tr}}=16$K & 2.86 \Std{$\pm$ 0.03} & 3.40 \Std{$\pm$ 0.04} & 3.25 \Std{$\pm$ 0.04} & 3.17 \\
\addlinespace[2pt]

\multirow{3}{*}{VocabTrim} & $V_{\mathrm{tr}}=64$K & 3.46 \Std{$\pm$ 0.03} & 4.37 \Std{$\pm$ 0.04} & 3.81 \Std{$\pm$ 0.05} & 3.88 \\
 & $V_{\mathrm{tr}}=32$K & 3.41 \Std{$\pm$ 0.02} & 4.21 \Std{$\pm$ 0.04} & 3.79 \Std{$\pm$ 0.03} & 3.80 \\
 & $V_{\mathrm{tr}}=16$K & 3.22 \Std{$\pm$ 0.04} & 3.95 \Std{$\pm$ 0.02} & 3.58 \Std{$\pm$ 0.02} & 3.58 \\
\addlinespace[2pt]

BCL & $V_{\mathrm{tr}}=15.8$K & 3.25 \Std{$\pm$ 0.03} & 3.93 \Std{$\pm$ 0.03} & 3.66 \Std{$\pm$ 0.04} & 3.61 \\
\addlinespace[2pt]

\multirow{3}{*}{VocabTrim-T} & $V_{\mathrm{tr}}=64$K & 3.48 \Std{$\pm$ 0.05} & 4.39 \Std{$\pm$ 0.07} & 3.83 \Std{$\pm$ 0.05} & 3.90 \\
 & $V_{\mathrm{tr}}=32$K & 3.41 \Std{$\pm$ 0.02} & 4.19 \Std{$\pm$ 0.03} & 3.74 \Std{$\pm$ 0.06} & 3.78 \\
 & $V_{\mathrm{tr}}=16$K & 3.23 \Std{$\pm$ 0.00} & 3.92 \Std{$\pm$ 0.04} & 3.60 \Std{$\pm$ 0.04} & 3.58 \\
\addlinespace[2pt]

\multirow{3}{*}{SpecVocab} & $r=d/4$ & 3.51 \Std{$\pm$ 0.04} & 4.43 \Std{$\pm$ 0.03} & 3.87 \Std{$\pm$ 0.04} & 3.94 \\
 & $r=d/8$ & 3.52 \Std{$\pm$ 0.05} & 4.44 \Std{$\pm$ 0.02} & 3.94 \Std{$\pm$ 0.03} & 3.97 \\
 & $r=d/16$ & 3.51 \Std{$\pm$ 0.02} & 4.45 \Std{$\pm$ 0.05} & 3.90 \Std{$\pm$ 0.05} & 3.95 \\
\midrule
\multirow{3}{*}{SlimSpec} & $r=d/4$ & 3.46 \Std{$\pm$ 0.03} & 4.45 \Std{$\pm$ 0.03} & 3.91 \Std{$\pm$ 0.05} & 3.94 \\
 & $r=d/8$ & 3.41 \Std{$\pm$ 0.02} & 4.47 \Std{$\pm$ 0.05} & 3.88 \Std{$\pm$ 0.04} & 3.92 \\
 & $r=d/16$ & 3.35 \Std{$\pm$ 0.02} & 4.32 \Std{$\pm$ 0.05} & 3.84 \Std{$\pm$ 0.05} & 3.84 \\
\bottomrule
\end{tabular}
\end{table}

\clearpage
\begin{table}[htbp]
\centering
\caption{Speedup for GPT-OSS-20B, temperature $=0$, batch size $=1$.}
\label{tab:gpt-oss-20b-t0-bs1-speedup}

\setlength{\tabcolsep}{9pt}
\renewcommand{\arraystretch}{1.1}
\begin{tabular}{@{}ll ccc>{\columncolor{gray!12}}c@{}}
\toprule
\textbf{Method} & \textbf{Config} & \textbf{MT-Bench} & \textbf{Humaneval} & \textbf{GSM8K} & \textbf{Avg} \\
\midrule
Full Vocab & -- & 1.43 \Std{$\pm$ 0.00} & 1.41 \Std{$\pm$ 0.00} & 1.63 \Std{$\pm$ 0.01} & 1.49 \\
\addlinespace[2pt]

\multirow{2}{*}{FR-Spec} & $V_{\mathrm{tr}}=64$K & 1.64 \Std{$\pm$ 0.01} & 1.57 \Std{$\pm$ 0.01} & 1.82 \Std{$\pm$ 0.00} & 1.68 \\
 & $V_{\mathrm{tr}}=32$K & 1.52 \Std{$\pm$ 0.01} & 1.50 \Std{$\pm$ 0.01} & 1.71 \Std{$\pm$ 0.02} & 1.57 \\
\addlinespace[2pt]

\multirow{2}{*}{VocabTrim} & $V_{\mathrm{tr}}=64$K & 1.71 \Std{$\pm$ 0.01} & 1.66 \Std{$\pm$ 0.01} & 1.89 \Std{$\pm$ 0.01} & 1.75 \\
 & $V_{\mathrm{tr}}=32$K & 1.64 \Std{$\pm$ 0.01} & 1.60 \Std{$\pm$ 0.01} & 1.87 \Std{$\pm$ 0.01} & 1.71 \\
\addlinespace[2pt]

BCL & $V_{\mathrm{tr}}=81.7$K & 1.72 \Std{$\pm$ 0.02} & {\bf 1.78} \Std{$\pm$ 0.02} & 1.99 \Std{$\pm$ 0.01} & 1.83 \\
\addlinespace[2pt]

\multirow{2}{*}{VocabTrim-T} & $V_{\mathrm{tr}}=64$K & 1.72 \Std{$\pm$ 0.01} & 1.67 \Std{$\pm$ 0.00} & 1.95 \Std{$\pm$ 0.00} & 1.78 \\
 & $V_{\mathrm{tr}}=32$K & 1.76 \Std{$\pm$ 0.00} & 1.71 \Std{$\pm$ 0.01} & 1.96 \Std{$\pm$ 0.00} & 1.81 \\
\addlinespace[2pt]

SpecVocab & $r=d/8$ & 1.74 \Std{$\pm$ 0.00} & 1.71 \Std{$\pm$ 0.01} & 1.98 \Std{$\pm$ 0.01} & 1.81 \\
\midrule
\multirow{3}{*}{SlimSpec} & $r=d/4$ & {\bf 1.77} \Std{$\pm$ 0.01} & 1.75 \Std{$\pm$ 0.01} & 2.06 \Std{$\pm$ 0.01} & 1.86 \\
 & $r=d/8$ & 1.76 \Std{$\pm$ 0.02} & 1.76 \Std{$\pm$ 0.01} & {\bf 2.11} \Std{$\pm$ 0.01} & {\bf 1.88} \\
 & $r=d/16$ & 1.76 \Std{$\pm$ 0.01} & 1.74 \Std{$\pm$ 0.01} & 2.05 \Std{$\pm$ 0.00} & 1.85 \\
\bottomrule
\end{tabular}
\end{table}

\begin{table}[htbp]
\centering
\caption{Average acceptance length $\tau$ for GPT-OSS-20B, temperature $=0$, batch size $=1$.}
\label{tab:gpt-oss-20b-t0-bs1-tau}

\setlength{\tabcolsep}{9pt}
\renewcommand{\arraystretch}{1.1}
\begin{tabular}{@{}ll ccc>{\columncolor{gray!12}}c@{}}
\toprule
\textbf{Method} & \textbf{Config} & \textbf{MT-Bench} & \textbf{Humaneval} & \textbf{GSM8K} & \textbf{Avg} \\
\midrule
Full Vocab & -- & 3.30 \Std{$\pm$ 0.03} & 3.36 \Std{$\pm$ 0.00} & 3.89 \Std{$\pm$ 0.02} & 3.52 \\
\addlinespace[2pt]

\multirow{2}{*}{FR-Spec} & $V_{\mathrm{tr}}=64$K & 3.15 \Std{$\pm$ 0.02} & 3.10 \Std{$\pm$ 0.01} & 3.62 \Std{$\pm$ 0.01} & 3.29 \\
 & $V_{\mathrm{tr}}=32$K & 2.92 \Std{$\pm$ 0.03} & 2.95 \Std{$\pm$ 0.00} & 3.38 \Std{$\pm$ 0.00} & 3.08 \\
\addlinespace[2pt]

\multirow{2}{*}{VocabTrim} & $V_{\mathrm{tr}}=64$K & 3.27 \Std{$\pm$ 0.00} & 3.25 \Std{$\pm$ 0.01} & 3.71 \Std{$\pm$ 0.01} & 3.41 \\
 & $V_{\mathrm{tr}}=32$K & 3.16 \Std{$\pm$ 0.01} & 3.12 \Std{$\pm$ 0.01} & 3.63 \Std{$\pm$ 0.01} & 3.30 \\
\addlinespace[2pt]

BCL & $V_{\mathrm{tr}}=81.7$K & 3.34 \Std{$\pm$ 0.03} & 3.41 \Std{$\pm$ 0.01} & 3.95 \Std{$\pm$ 0.00} & 3.57 \\
\addlinespace[2pt]

\multirow{2}{*}{VocabTrim-T} & $V_{\mathrm{tr}}=64$K & 3.34 \Std{$\pm$ 0.00} & 3.35 \Std{$\pm$ 0.00} & 3.95 \Std{$\pm$ 0.01} & 3.55 \\
 & $V_{\mathrm{tr}}=32$K & 3.20 \Std{$\pm$ 0.01} & 3.19 \Std{$\pm$ 0.02} & 3.69 \Std{$\pm$ 0.01} & 3.36 \\
\addlinespace[2pt]

SpecVocab & $r=d/8$ & 3.37 \Std{$\pm$ 0.00} & 3.40 \Std{$\pm$ 0.02} & 3.99 \Std{$\pm$ 0.01} & 3.59 \\
\midrule
\multirow{3}{*}{SlimSpec} & $r=d/4$ & 3.34 \Std{$\pm$ 0.01} & 3.39 \Std{$\pm$ 0.02} & 4.03 \Std{$\pm$ 0.00} & 3.59 \\
 & $r=d/8$ & 3.18 \Std{$\pm$ 0.02} & 3.30 \Std{$\pm$ 0.01} & 3.99 \Std{$\pm$ 0.01} & 3.49 \\
 & $r=d/16$ & 3.20 \Std{$\pm$ 0.01} & 3.25 \Std{$\pm$ 0.02} & 3.90 \Std{$\pm$ 0.00} & 3.45 \\
\bottomrule
\end{tabular}
\end{table}

\clearpage
\begin{table}[htbp]
\centering
\caption{Speedup for GPT-OSS-20B, temperature $=0$, batch size $=64$.}
\label{tab:gpt-oss-20b-t0-bs64-speedup}

\setlength{\tabcolsep}{9pt}
\renewcommand{\arraystretch}{1.1}
\begin{tabular}{@{}ll ccc>{\columncolor{gray!12}}c@{}}
\toprule
\textbf{Method} & \textbf{Config} & \textbf{MT-Bench} & \textbf{Humaneval} & \textbf{GSM8K} & \textbf{Avg} \\
\midrule
Full Vocab & -- & 1.36 \Std{$\pm$ 0.03} & 1.17 \Std{$\pm$ 0.02} & 1.38 \Std{$\pm$ 0.01} & 1.30 \\
\addlinespace[2pt]

\multirow{2}{*}{FR-Spec} & $V_{\mathrm{tr}}=64$K & 1.38 \Std{$\pm$ 0.03} & 1.15 \Std{$\pm$ 0.03} & 1.42 \Std{$\pm$ 0.00} & 1.32 \\
 & $V_{\mathrm{tr}}=32$K & 1.35 \Std{$\pm$ 0.01} & 1.15 \Std{$\pm$ 0.02} & 1.40 \Std{$\pm$ 0.02} & 1.30 \\
\addlinespace[2pt]

\multirow{2}{*}{VocabTrim} & $V_{\mathrm{tr}}=64$K & {\bf 1.52} \Std{$\pm$ 0.01} & 1.29 \Std{$\pm$ 0.03} & 1.48 \Std{$\pm$ 0.02} & 1.43 \\
 & $V_{\mathrm{tr}}=32$K & 1.50 \Std{$\pm$ 0.03} & 1.27 \Std{$\pm$ 0.00} & 1.49 \Std{$\pm$ 0.00} & 1.42 \\
\addlinespace[2pt]

BCL & $V_{\mathrm{tr}}=81.7$K & 1.49 \Std{$\pm$ 0.02} & {\bf 1.34} \Std{$\pm$ 0.04} & {\bf 1.56} \Std{$\pm$ 0.03} & 1.46 \\
\addlinespace[2pt]

\multirow{2}{*}{VocabTrim-T} & $V_{\mathrm{tr}}=64$K & 1.50 \Std{$\pm$ 0.04} & 1.27 \Std{$\pm$ 0.01} & 1.52 \Std{$\pm$ 0.03} & 1.43 \\
 & $V_{\mathrm{tr}}=32$K & 1.41 \Std{$\pm$ 0.02} & 1.20 \Std{$\pm$ 0.02} & 1.44 \Std{$\pm$ 0.06} & 1.35 \\
\addlinespace[2pt]

SpecVocab & $r=d/8$ & 1.43 \Std{$\pm$ 0.02} & 1.18 \Std{$\pm$ 0.03} & 1.45 \Std{$\pm$ 0.02} & 1.35 \\
\midrule
\multirow{3}{*}{SlimSpec} & $r=d/4$ & 1.44 \Std{$\pm$ 0.01} & 1.23 \Std{$\pm$ 0.06} & 1.50 \Std{$\pm$ 0.02} & 1.39 \\
 & $r=d/8$ & {\bf 1.52} \Std{$\pm$ 0.03} & {\bf 1.34} \Std{$\pm$ 0.02} & {\bf 1.56} \Std{$\pm$ 0.04} & {\bf 1.47} \\
 & $r=d/16$ & 1.45 \Std{$\pm$ 0.03} & 1.20 \Std{$\pm$ 0.04} & 1.48 \Std{$\pm$ 0.05} & 1.38 \\
\bottomrule
\end{tabular}
\end{table}

\begin{table}[htbp]
\centering
\caption{Average acceptance length $\tau$ for GPT-OSS-20B, temperature $=0$, batch size $=64$.}
\label{tab:gpt-oss-20b-t0-bs64-tau}

\setlength{\tabcolsep}{9pt}
\renewcommand{\arraystretch}{1.1}
\begin{tabular}{@{}ll ccc>{\columncolor{gray!12}}c@{}}
\toprule
\textbf{Method} & \textbf{Config} & \textbf{MT-Bench} & \textbf{Humaneval} & \textbf{GSM8K} & \textbf{Avg} \\
\midrule
Full Vocab & -- & 3.33 \Std{$\pm$ 0.03} & 3.36 \Std{$\pm$ 0.02} & 3.91 \Std{$\pm$ 0.02} & 3.53 \\
\addlinespace[2pt]

\multirow{2}{*}{FR-Spec} & $V_{\mathrm{tr}}=64$K & 3.10 \Std{$\pm$ 0.04} & 3.10 \Std{$\pm$ 0.01} & 3.62 \Std{$\pm$ 0.01} & 3.27 \\
 & $V_{\mathrm{tr}}=32$K & 2.92 \Std{$\pm$ 0.02} & 2.95 \Std{$\pm$ 0.01} & 3.39 \Std{$\pm$ 0.00} & 3.09 \\
\addlinespace[2pt]

\multirow{2}{*}{VocabTrim} & $V_{\mathrm{tr}}=64$K & 3.26 \Std{$\pm$ 0.02} & 3.26 \Std{$\pm$ 0.02} & 3.72 \Std{$\pm$ 0.01} & 3.41 \\
 & $V_{\mathrm{tr}}=32$K & 3.13 \Std{$\pm$ 0.01} & 3.12 \Std{$\pm$ 0.02} & 3.61 \Std{$\pm$ 0.02} & 3.29 \\
\addlinespace[2pt]

BCL & $V_{\mathrm{tr}}=81.7$K & 3.38 \Std{$\pm$ 0.02} & 3.39 \Std{$\pm$ 0.01} & 3.94 \Std{$\pm$ 0.02} & 3.57 \\
\addlinespace[2pt]

\multirow{2}{*}{VocabTrim-T} & $V_{\mathrm{tr}}=64$K & 3.32 \Std{$\pm$ 0.01} & 3.35 \Std{$\pm$ 0.02} & 3.96 \Std{$\pm$ 0.02} & 3.54 \\
 & $V_{\mathrm{tr}}=32$K & 3.19 \Std{$\pm$ 0.01} & 3.20 \Std{$\pm$ 0.02} & 3.69 \Std{$\pm$ 0.02} & 3.36 \\
\addlinespace[2pt]

SpecVocab & $r=d/8$ & 3.37 \Std{$\pm$ 0.03} & 3.39 \Std{$\pm$ 0.02} & 3.99 \Std{$\pm$ 0.02} & 3.58 \\
\midrule
\multirow{3}{*}{SlimSpec} & $r=d/4$ & 3.38 \Std{$\pm$ 0.02} & 3.38 \Std{$\pm$ 0.01} & 4.03 \Std{$\pm$ 0.02} & 3.60 \\
 & $r=d/8$ & 3.29 \Std{$\pm$ 0.02} & 3.34 \Std{$\pm$ 0.02} & 3.95 \Std{$\pm$ 0.03} & 3.53 \\
 & $r=d/16$ & 3.21 \Std{$\pm$ 0.02} & 3.24 \Std{$\pm$ 0.02} & 3.89 \Std{$\pm$ 0.01} & 3.45 \\
\bottomrule
\end{tabular}
\end{table}

\clearpage
\begin{table}[htbp]
\centering
\caption{Speedup for GPT-OSS-20B, temperature $=1$, batch size $=1$.}
\label{tab:gpt-oss-20b-t1-bs1-speedup}

\setlength{\tabcolsep}{9pt}
\renewcommand{\arraystretch}{1.1}
\begin{tabular}{@{}ll ccc>{\columncolor{gray!12}}c@{}}
\toprule
\textbf{Method} & \textbf{Config} & \textbf{MT-Bench} & \textbf{Humaneval} & \textbf{GSM8K} & \textbf{Avg} \\
\midrule
Full Vocab & -- & 1.24 \Std{$\pm$ 0.01} & 1.28 \Std{$\pm$ 0.01} & 1.47 \Std{$\pm$ 0.01} & 1.33 \\
\addlinespace[2pt]

\multirow{2}{*}{FR-Spec} & $V_{\mathrm{tr}}=64$K & 1.17 \Std{$\pm$ 0.01} & 1.23 \Std{$\pm$ 0.02} & 1.38 \Std{$\pm$ 0.05} & 1.26 \\
 & $V_{\mathrm{tr}}=32$K & 1.07 \Std{$\pm$ 0.01} & 1.17 \Std{$\pm$ 0.01} & 1.28 \Std{$\pm$ 0.01} & 1.17 \\
\addlinespace[2pt]

\multirow{2}{*}{VocabTrim} & $V_{\mathrm{tr}}=64$K & 1.21 \Std{$\pm$ 0.04} & 1.33 \Std{$\pm$ 0.02} & 1.42 \Std{$\pm$ 0.02} & 1.32 \\
 & $V_{\mathrm{tr}}=32$K & 1.19 \Std{$\pm$ 0.01} & 1.20 \Std{$\pm$ 0.02} & 1.36 \Std{$\pm$ 0.04} & 1.25 \\
\addlinespace[2pt]

BCL & $V_{\mathrm{tr}}=81.7$K & 1.29 \Std{$\pm$ 0.03} & 1.35 \Std{$\pm$ 0.03} & 1.57 \Std{$\pm$ 0.02} & 1.40 \\
\addlinespace[2pt]

\multirow{2}{*}{VocabTrim-T} & $V_{\mathrm{tr}}=64$K & 1.28 \Std{$\pm$ 0.01} & 1.25 \Std{$\pm$ 0.02} & 1.48 \Std{$\pm$ 0.04} & 1.34 \\
 & $V_{\mathrm{tr}}=32$K & 1.38 \Std{$\pm$ 0.01} & 1.43 \Std{$\pm$ 0.01} & 1.68 \Std{$\pm$ 0.02} & 1.49 \\
\addlinespace[2pt]

SpecVocab & $r=d/8$ & 1.39 \Std{$\pm$ 0.01} & 1.46 \Std{$\pm$ 0.02} & 1.71 \Std{$\pm$ 0.01} & 1.52 \\
\midrule
\multirow{3}{*}{SlimSpec} & $r=d/4$ & {\bf 1.43} \Std{$\pm$ 0.02} & {\bf 1.52} \Std{$\pm$ 0.01} & {\bf 1.74} \Std{$\pm$ 0.01} & {\bf 1.56} \\
 & $r=d/8$ & 1.40 \Std{$\pm$ 0.02} & 1.47 \Std{$\pm$ 0.03} & 1.72 \Std{$\pm$ 0.02} & 1.53 \\
 & $r=d/16$ & 1.32 \Std{$\pm$ 0.01} & 1.38 \Std{$\pm$ 0.01} & 1.64 \Std{$\pm$ 0.02} & 1.44 \\
\bottomrule
\end{tabular}
\end{table}

\begin{table}[htbp]
\centering
\caption{Average acceptance length $\tau$ for GPT-OSS-20B, temperature $=1$, batch size $=1$.}
\label{tab:gpt-oss-20b-t1-bs1-tau}

\setlength{\tabcolsep}{9pt}
\renewcommand{\arraystretch}{1.1}
\begin{tabular}{@{}ll ccc>{\columncolor{gray!12}}c@{}}
\toprule
\textbf{Method} & \textbf{Config} & \textbf{MT-Bench} & \textbf{Humaneval} & \textbf{GSM8K} & \textbf{Avg} \\
\midrule
Full Vocab & -- & 2.96 \Std{$\pm$ 0.02} & 3.08 \Std{$\pm$ 0.02} & 3.44 \Std{$\pm$ 0.03} & 3.16 \\
\addlinespace[2pt]

\multirow{2}{*}{FR-Spec} & $V_{\mathrm{tr}}=64$K & 2.78 \Std{$\pm$ 0.03} & 2.89 \Std{$\pm$ 0.01} & 3.22 \Std{$\pm$ 0.02} & 2.96 \\
 & $V_{\mathrm{tr}}=32$K & 2.62 \Std{$\pm$ 0.02} & 2.76 \Std{$\pm$ 0.02} & 3.04 \Std{$\pm$ 0.01} & 2.81 \\
\addlinespace[2pt]

\multirow{2}{*}{VocabTrim} & $V_{\mathrm{tr}}=64$K & 2.89 \Std{$\pm$ 0.02} & 2.99 \Std{$\pm$ 0.03} & 3.30 \Std{$\pm$ 0.02} & 3.06 \\
 & $V_{\mathrm{tr}}=32$K & 2.80 \Std{$\pm$ 0.01} & 2.88 \Std{$\pm$ 0.02} & 3.21 \Std{$\pm$ 0.02} & 2.96 \\
\addlinespace[2pt]

BCL & $V_{\mathrm{tr}}=81.7$K & 3.08 \Std{$\pm$ 0.01} & 3.22 \Std{$\pm$ 0.02} & 3.60 \Std{$\pm$ 0.03} & 3.30 \\
\addlinespace[2pt]

\multirow{2}{*}{VocabTrim-T} & $V_{\mathrm{tr}}=64$K & 2.95 \Std{$\pm$ 0.03} & 3.08 \Std{$\pm$ 0.04} & 3.47 \Std{$\pm$ 0.02} & 3.17 \\
 & $V_{\mathrm{tr}}=32$K & 2.87 \Std{$\pm$ 0.03} & 2.99 \Std{$\pm$ 0.01} & 3.32 \Std{$\pm$ 0.01} & 3.06 \\
\addlinespace[2pt]

SpecVocab & $r=d/8$ & 2.93 \Std{$\pm$ 0.02} & 3.13 \Std{$\pm$ 0.02} & 3.50 \Std{$\pm$ 0.01} & 3.19 \\
\midrule
\multirow{3}{*}{SlimSpec} & $r=d/4$ & 2.96 \Std{$\pm$ 0.02} & 3.16 \Std{$\pm$ 0.02} & 3.54 \Std{$\pm$ 0.02} & 3.22 \\
 & $r=d/8$ & 2.83 \Std{$\pm$ 0.03} & 3.02 \Std{$\pm$ 0.02} & 3.42 \Std{$\pm$ 0.02} & 3.09 \\
 & $r=d/16$ & 2.76 \Std{$\pm$ 0.03} & 2.92 \Std{$\pm$ 0.01} & 3.35 \Std{$\pm$ 0.01} & 3.01 \\
\bottomrule
\end{tabular}
\end{table}

\clearpage
\begin{table}[htbp]
\centering
\caption{Speedup for GPT-OSS-20B, temperature $=1$, batch size $=64$.}
\label{tab:gpt-oss-20b-t1-bs64-speedup}

\setlength{\tabcolsep}{9pt}
\renewcommand{\arraystretch}{1.1}
\begin{tabular}{@{}ll ccc>{\columncolor{gray!12}}c@{}}
\toprule
\textbf{Method} & \textbf{Config} & \textbf{MT-Bench} & \textbf{Humaneval} & \textbf{GSM8K} & \textbf{Avg} \\
\midrule
Full Vocab & -- & 1.25 \Std{$\pm$ 0.02} & 1.17 \Std{$\pm$ 0.01} & 1.28 \Std{$\pm$ 0.01} & 1.23 \\
\addlinespace[2pt]

\multirow{2}{*}{FR-Spec} & $V_{\mathrm{tr}}=64$K & 1.31 \Std{$\pm$ 0.02} & 1.15 \Std{$\pm$ 0.03} & 1.30 \Std{$\pm$ 0.02} & 1.25 \\
 & $V_{\mathrm{tr}}=32$K & 1.26 \Std{$\pm$ 0.04} & 1.14 \Std{$\pm$ 0.01} & 1.26 \Std{$\pm$ 0.02} & 1.22 \\
\addlinespace[2pt]

\multirow{2}{*}{VocabTrim} & $V_{\mathrm{tr}}=64$K & 1.30 \Std{$\pm$ 0.04} & 1.24 \Std{$\pm$ 0.02} & 1.33 \Std{$\pm$ 0.02} & 1.29 \\
 & $V_{\mathrm{tr}}=32$K & 1.28 \Std{$\pm$ 0.05} & 1.25 \Std{$\pm$ 0.03} & 1.32 \Std{$\pm$ 0.02} & 1.28 \\
\addlinespace[2pt]

BCL & $V_{\mathrm{tr}}=81.7$K & 1.30 \Std{$\pm$ 0.04} & {\bf 1.28} \Std{$\pm$ 0.02} & {\bf 1.38} \Std{$\pm$ 0.03} & {\bf 1.32} \\
\addlinespace[2pt]

\multirow{2}{*}{VocabTrim-T} & $V_{\mathrm{tr}}=64$K & {\bf 1.34} \Std{$\pm$ 0.04} & 1.26 \Std{$\pm$ 0.04} & 1.36 \Std{$\pm$ 0.02} & {\bf 1.32} \\
 & $V_{\mathrm{tr}}=32$K & 1.29 \Std{$\pm$ 0.04} & 1.15 \Std{$\pm$ 0.04} & 1.36 \Std{$\pm$ 0.02} & 1.27 \\
\addlinespace[2pt]

SpecVocab & $r=d/8$ & 1.29 \Std{$\pm$ 0.02} & 1.18 \Std{$\pm$ 0.05} & 1.30 \Std{$\pm$ 0.08} & 1.26 \\
\midrule
\multirow{3}{*}{SlimSpec} & $r=d/4$ & {\bf 1.34} \Std{$\pm$ 0.03} & 1.20 \Std{$\pm$ 0.02} & 1.36 \Std{$\pm$ 0.07} & 1.30 \\
 & $r=d/8$ & 1.32 \Std{$\pm$ 0.03} & {\bf 1.28} \Std{$\pm$ 0.02} & 1.35 \Std{$\pm$ 0.06} & {\bf 1.32} \\
 & $r=d/16$ & 1.17 \Std{$\pm$ 0.26} & 1.17 \Std{$\pm$ 0.02} & 1.33 \Std{$\pm$ 0.06} & 1.22 \\
\bottomrule
\end{tabular}
\end{table}

\begin{table}[htbp]
\centering
\caption{Average acceptance length $\tau$ for GPT-OSS-20B, temperature $=1$, batch size $=64$.}
\label{tab:gpt-oss-20b-t1-bs64-tau}

\setlength{\tabcolsep}{9pt}
\renewcommand{\arraystretch}{1.1}
\begin{tabular}{@{}ll ccc>{\columncolor{gray!12}}c@{}}
\toprule
\textbf{Method} & \textbf{Config} & \textbf{MT-Bench} & \textbf{Humaneval} & \textbf{GSM8K} & \textbf{Avg} \\
\midrule
Full Vocab & -- & 2.95 \Std{$\pm$ 0.02} & 3.08 \Std{$\pm$ 0.03} & 3.41 \Std{$\pm$ 0.02} & 3.15 \\
\addlinespace[2pt]

\multirow{2}{*}{FR-Spec} & $V_{\mathrm{tr}}=64$K & 2.80 \Std{$\pm$ 0.02} & 2.88 \Std{$\pm$ 0.03} & 3.22 \Std{$\pm$ 0.02} & 2.97 \\
 & $V_{\mathrm{tr}}=32$K & 2.62 \Std{$\pm$ 0.01} & 2.75 \Std{$\pm$ 0.02} & 3.03 \Std{$\pm$ 0.02} & 2.80 \\
\addlinespace[2pt]

\multirow{2}{*}{VocabTrim} & $V_{\mathrm{tr}}=64$K & 2.86 \Std{$\pm$ 0.01} & 3.00 \Std{$\pm$ 0.02} & 3.29 \Std{$\pm$ 0.01} & 3.05 \\
 & $V_{\mathrm{tr}}=32$K & 2.79 \Std{$\pm$ 0.03} & 2.91 \Std{$\pm$ 0.02} & 3.20 \Std{$\pm$ 0.03} & 2.97 \\
\addlinespace[2pt]

BCL & $V_{\mathrm{tr}}=81.7$K & 3.10 \Std{$\pm$ 0.03} & 3.23 \Std{$\pm$ 0.00} & 3.63 \Std{$\pm$ 0.02} & 3.32 \\
\addlinespace[2pt]

\multirow{2}{*}{VocabTrim-T} & $V_{\mathrm{tr}}=64$K & 2.94 \Std{$\pm$ 0.01} & 3.06 \Std{$\pm$ 0.03} & 3.47 \Std{$\pm$ 0.02} & 3.16 \\
 & $V_{\mathrm{tr}}=32$K & 2.86 \Std{$\pm$ 0.02} & 2.99 \Std{$\pm$ 0.03} & 3.32 \Std{$\pm$ 0.02} & 3.06 \\
\addlinespace[2pt]

SpecVocab & $r=d/8$ & 2.95 \Std{$\pm$ 0.03} & 3.13 \Std{$\pm$ 0.02} & 3.49 \Std{$\pm$ 0.03} & 3.19 \\
\midrule
\multirow{3}{*}{SlimSpec} & $r=d/4$ & 2.96 \Std{$\pm$ 0.02} & 3.15 \Std{$\pm$ 0.02} & 3.55 \Std{$\pm$ 0.01} & 3.22 \\
 & $r=d/8$ & 2.84 \Std{$\pm$ 0.03} & 3.02 \Std{$\pm$ 0.05} & 3.40 \Std{$\pm$ 0.03} & 3.09 \\
 & $r=d/16$ & 2.75 \Std{$\pm$ 0.02} & 2.92 \Std{$\pm$ 0.02} & 3.35 \Std{$\pm$ 0.01} & 3.01 \\
\bottomrule
\end{tabular}
\end{table}

\clearpage
\begin{table}[htbp]
\centering
\caption{Speedup for Qwen3-30B-A3B, temperature $=0$, batch size $=1$.}
\label{tab:qwen3-30b-t0-bs1-speedup}

\setlength{\tabcolsep}{9pt}
\renewcommand{\arraystretch}{1.1}
\begin{tabular}{@{}ll ccc>{\columncolor{gray!12}}c@{}}
\toprule
\textbf{Method} & \textbf{Config} & \textbf{MT-Bench} & \textbf{Humaneval} & \textbf{GSM8K} & \textbf{Avg} \\
\midrule
Full Vocab & -- & 1.55 \Std{$\pm$ 0.02} & 2.07 \Std{$\pm$ 0.01} & 2.12 \Std{$\pm$ 0.01} & 1.91 \\
\addlinespace[2pt]

\multirow{2}{*}{FR-Spec} & $V_{\mathrm{tr}}=64$K & 1.59 \Std{$\pm$ 0.01} & 2.15 \Std{$\pm$ 0.02} & 2.13 \Std{$\pm$ 0.01} & 1.96 \\
 & $V_{\mathrm{tr}}=32$K & 1.51 \Std{$\pm$ 0.00} & 1.95 \Std{$\pm$ 0.01} & 1.92 \Std{$\pm$ 0.01} & 1.79 \\
\addlinespace[2pt]

\multirow{2}{*}{VocabTrim} & $V_{\mathrm{tr}}=64$K & 1.65 \Std{$\pm$ 0.01} & 2.19 \Std{$\pm$ 0.02} & 2.22 \Std{$\pm$ 0.01} & 2.02 \\
 & $V_{\mathrm{tr}}=32$K & 1.66 \Std{$\pm$ 0.01} & 2.18 \Std{$\pm$ 0.01} & 2.22 \Std{$\pm$ 0.02} & 2.02 \\
\addlinespace[2pt]

BCL & $V_{\mathrm{tr}}=66.9$K & 1.62 \Std{$\pm$ 0.00} & 2.08 \Std{$\pm$ 0.01} & 2.16 \Std{$\pm$ 0.01} & 1.95 \\
\addlinespace[2pt]

\multirow{2}{*}{VocabTrim-T} & $V_{\mathrm{tr}}=64$K & 1.66 \Std{$\pm$ 0.01} & 2.15 \Std{$\pm$ 0.01} & 2.23 \Std{$\pm$ 0.01} & 2.01 \\
 & $V_{\mathrm{tr}}=32$K & 1.66 \Std{$\pm$ 0.01} & 2.15 \Std{$\pm$ 0.02} & 2.22 \Std{$\pm$ 0.01} & 2.01 \\
\addlinespace[2pt]

SpecVocab & $r=d/8$ & 1.66 \Std{$\pm$ 0.01} & 2.19 \Std{$\pm$ 0.01} & 2.24 \Std{$\pm$ 0.01} & 2.03 \\
\midrule
\multirow{3}{*}{SlimSpec} & $r=d/4$ & {\bf 1.68} \Std{$\pm$ 0.03} & {\bf 2.22} \Std{$\pm$ 0.03} & {\bf 2.27} \Std{$\pm$ 0.02} & {\bf 2.05} \\
 & $r=d/8$ & 1.66 \Std{$\pm$ 0.00} & {\bf 2.22} \Std{$\pm$ 0.02} & {\bf 2.27} \Std{$\pm$ 0.02} & {\bf 2.05} \\
 & $r=d/16$ & 1.61 \Std{$\pm$ 0.00} & 2.19 \Std{$\pm$ 0.01} & 2.23 \Std{$\pm$ 0.00} & 2.01 \\
\bottomrule
\end{tabular}
\end{table}

\begin{table}[htbp]
\centering
\caption{Average acceptance length $\tau$ for Qwen3-30B-A3B, temperature $=0$, batch size $=1$.}
\label{tab:qwen3-30b-t0-bs1-tau}

\setlength{\tabcolsep}{9pt}
\renewcommand{\arraystretch}{1.1}
\begin{tabular}{@{}ll ccc>{\columncolor{gray!12}}c@{}}
\toprule
\textbf{Method} & \textbf{Config} & \textbf{MT-Bench} & \textbf{Humaneval} & \textbf{GSM8K} & \textbf{Avg} \\
\midrule
Full Vocab & -- & 3.12 \Std{$\pm$ 0.00} & 4.37 \Std{$\pm$ 0.01} & 4.52 \Std{$\pm$ 0.00} & 4.00 \\
\addlinespace[2pt]

\multirow{2}{*}{FR-Spec} & $V_{\mathrm{tr}}=64$K & 2.99 \Std{$\pm$ 0.01} & 4.26 \Std{$\pm$ 0.01} & 4.32 \Std{$\pm$ 0.01} & 3.86 \\
 & $V_{\mathrm{tr}}=32$K & 2.76 \Std{$\pm$ 0.01} & 3.78 \Std{$\pm$ 0.00} & 3.78 \Std{$\pm$ 0.01} & 3.44 \\
\addlinespace[2pt]

\multirow{2}{*}{VocabTrim} & $V_{\mathrm{tr}}=64$K & 3.11 \Std{$\pm$ 0.00} & 4.35 \Std{$\pm$ 0.00} & 4.51 \Std{$\pm$ 0.00} & 3.99 \\
 & $V_{\mathrm{tr}}=32$K & 3.06 \Std{$\pm$ 0.00} & 4.22 \Std{$\pm$ 0.01} & 4.40 \Std{$\pm$ 0.00} & 3.89 \\
\addlinespace[2pt]

BCL & $V_{\mathrm{tr}}=66.9$K & 2.96 \Std{$\pm$ 0.00} & 3.98 \Std{$\pm$ 0.01} & 4.25 \Std{$\pm$ 0.00} & 3.73 \\
\addlinespace[2pt]

\multirow{2}{*}{VocabTrim-T} & $V_{\mathrm{tr}}=64$K & 3.12 \Std{$\pm$ 0.00} & 4.29 \Std{$\pm$ 0.01} & 4.52 \Std{$\pm$ 0.00} & 3.98 \\
 & $V_{\mathrm{tr}}=32$K & 3.05 \Std{$\pm$ 0.00} & 4.17 \Std{$\pm$ 0.00} & 4.40 \Std{$\pm$ 0.00} & 3.87 \\
\addlinespace[2pt]

SpecVocab & $r=d/8$ & 3.13 \Std{$\pm$ 0.00} & 4.36 \Std{$\pm$ 0.01} & 4.54 \Std{$\pm$ 0.01} & 4.01 \\
\midrule
\multirow{3}{*}{SlimSpec} & $r=d/4$ & 3.09 \Std{$\pm$ 0.00} & 4.31 \Std{$\pm$ 0.00} & 4.52 \Std{$\pm$ 0.01} & 3.97 \\
 & $r=d/8$ & 3.00 \Std{$\pm$ 0.01} & 4.26 \Std{$\pm$ 0.00} & 4.47 \Std{$\pm$ 0.00} & 3.91 \\
 & $r=d/16$ & 2.92 \Std{$\pm$ 0.01} & 4.20 \Std{$\pm$ 0.00} & 4.37 \Std{$\pm$ 0.00} & 3.83 \\
\bottomrule
\end{tabular}
\end{table}

\clearpage
\begin{table}[htbp]
\centering
\caption{Speedup for Qwen3-30B-A3B, temperature $=0$, batch size $=64$.}
\label{tab:qwen3-30b-t0-bs64-speedup}

\setlength{\tabcolsep}{9pt}
\renewcommand{\arraystretch}{1.1}
\begin{tabular}{@{}ll ccc>{\columncolor{gray!12}}c@{}}
\toprule
\textbf{Method} & \textbf{Config} & \textbf{MT-Bench} & \textbf{Humaneval} & \textbf{GSM8K} & \textbf{Avg} \\
\midrule
Full Vocab & -- & 1.22 \Std{$\pm$ 0.02} & 1.32 \Std{$\pm$ 0.02} & 1.48 \Std{$\pm$ 0.02} & 1.34 \\
\addlinespace[2pt]

\multirow{2}{*}{FR-Spec} & $V_{\mathrm{tr}}=64$K & 1.27 \Std{$\pm$ 0.01} & 1.36 \Std{$\pm$ 0.03} & 1.48 \Std{$\pm$ 0.02} & 1.37 \\
 & $V_{\mathrm{tr}}=32$K & 1.24 \Std{$\pm$ 0.02} & 1.31 \Std{$\pm$ 0.03} & 1.40 \Std{$\pm$ 0.02} & 1.32 \\
\addlinespace[2pt]

\multirow{2}{*}{VocabTrim} & $V_{\mathrm{tr}}=64$K & 1.30 \Std{$\pm$ 0.02} & 1.36 \Std{$\pm$ 0.02} & 1.51 \Std{$\pm$ 0.02} & 1.39 \\
 & $V_{\mathrm{tr}}=32$K & 1.29 \Std{$\pm$ 0.02} & 1.35 \Std{$\pm$ 0.03} & 1.52 \Std{$\pm$ 0.02} & 1.39 \\
\addlinespace[2pt]

BCL & $V_{\mathrm{tr}}=66.9$K & {\bf 1.32} \Std{$\pm$ 0.03} & 1.35 \Std{$\pm$ 0.01} & 1.49 \Std{$\pm$ 0.03} & 1.39 \\
\addlinespace[2pt]

\multirow{2}{*}{VocabTrim-T} & $V_{\mathrm{tr}}=64$K & 1.31 \Std{$\pm$ 0.02} & 1.36 \Std{$\pm$ 0.02} & 1.53 \Std{$\pm$ 0.04} & {\bf 1.40} \\
 & $V_{\mathrm{tr}}=32$K & 1.30 \Std{$\pm$ 0.02} & 1.35 \Std{$\pm$ 0.02} & 1.50 \Std{$\pm$ 0.01} & 1.38 \\
\addlinespace[2pt]

SpecVocab & $r=d/8$ & 1.28 \Std{$\pm$ 0.03} & 1.34 \Std{$\pm$ 0.11} & 1.49 \Std{$\pm$ 0.02} & 1.37 \\
\midrule
\multirow{3}{*}{SlimSpec} & $r=d/4$ & 1.31 \Std{$\pm$ 0.04} & {\bf 1.37} \Std{$\pm$ 0.05} & 1.52 \Std{$\pm$ 0.04} & {\bf 1.40} \\
 & $r=d/8$ & 1.31 \Std{$\pm$ 0.02} & 1.36 \Std{$\pm$ 0.03} & {\bf 1.54} \Std{$\pm$ 0.03} & {\bf 1.40} \\
 & $r=d/16$ & 1.31 \Std{$\pm$ 0.04} & {\bf 1.37} \Std{$\pm$ 0.02} & 1.49 \Std{$\pm$ 0.04} & 1.39 \\
\bottomrule
\end{tabular}
\end{table}

\begin{table}[htbp]
\centering
\caption{Average acceptance length $\tau$ for Qwen3-30B-A3B, temperature $=0$, batch size $=64$.}
\label{tab:qwen3-30b-t0-bs64-tau}

\setlength{\tabcolsep}{9pt}
\renewcommand{\arraystretch}{1.1}
\begin{tabular}{@{}ll ccc>{\columncolor{gray!12}}c@{}}
\toprule
\textbf{Method} & \textbf{Config} & \textbf{MT-Bench} & \textbf{Humaneval} & \textbf{GSM8K} & \textbf{Avg} \\
\midrule
Full Vocab & -- & 3.12 \Std{$\pm$ 0.01} & 4.30 \Std{$\pm$ 0.04} & 4.50 \Std{$\pm$ 0.03} & 3.97 \\
\addlinespace[2pt]

\multirow{2}{*}{FR-Spec} & $V_{\mathrm{tr}}=64$K & 2.99 \Std{$\pm$ 0.01} & 4.17 \Std{$\pm$ 0.05} & 4.29 \Std{$\pm$ 0.01} & 3.82 \\
 & $V_{\mathrm{tr}}=32$K & 2.77 \Std{$\pm$ 0.01} & 3.69 \Std{$\pm$ 0.03} & 3.74 \Std{$\pm$ 0.02} & 3.40 \\
\addlinespace[2pt]

\multirow{2}{*}{VocabTrim} & $V_{\mathrm{tr}}=64$K & 3.12 \Std{$\pm$ 0.01} & 4.25 \Std{$\pm$ 0.04} & 4.47 \Std{$\pm$ 0.02} & 3.95 \\
 & $V_{\mathrm{tr}}=32$K & 3.07 \Std{$\pm$ 0.01} & 4.13 \Std{$\pm$ 0.03} & 4.37 \Std{$\pm$ 0.03} & 3.86 \\
\addlinespace[2pt]

BCL & $V_{\mathrm{tr}}=66.9$K & 2.93 \Std{$\pm$ 0.01} & 3.92 \Std{$\pm$ 0.01} & 4.26 \Std{$\pm$ 0.01} & 3.70 \\
\addlinespace[2pt]

\multirow{2}{*}{VocabTrim-T} & $V_{\mathrm{tr}}=64$K & 3.13 \Std{$\pm$ 0.01} & 4.24 \Std{$\pm$ 0.02} & 4.51 \Std{$\pm$ 0.01} & 3.96 \\
 & $V_{\mathrm{tr}}=32$K & 3.06 \Std{$\pm$ 0.01} & 4.12 \Std{$\pm$ 0.02} & 4.40 \Std{$\pm$ 0.01} & 3.86 \\
\addlinespace[2pt]

SpecVocab & $r=d/8$ & 3.14 \Std{$\pm$ 0.01} & 4.34 \Std{$\pm$ 0.02} & 4.56 \Std{$\pm$ 0.01} & 4.01 \\
\midrule
\multirow{3}{*}{SlimSpec} & $r=d/4$ & 3.07 \Std{$\pm$ 0.01} & 4.26 \Std{$\pm$ 0.04} & 4.50 \Std{$\pm$ 0.00} & 3.94 \\
 & $r=d/8$ & 3.01 \Std{$\pm$ 0.01} & 4.22 \Std{$\pm$ 0.02} & 4.45 \Std{$\pm$ 0.02} & 3.89 \\
 & $r=d/16$ & 2.92 \Std{$\pm$ 0.01} & 4.17 \Std{$\pm$ 0.02} & 4.36 \Std{$\pm$ 0.01} & 3.82 \\
\bottomrule
\end{tabular}
\end{table}

\clearpage
\begin{table}[htbp]
\centering
\caption{Speedup for Qwen3-30B-A3B, temperature $=1$, batch size $=1$.}
\label{tab:qwen3-30b-t1-bs1-speedup}

\setlength{\tabcolsep}{9pt}
\renewcommand{\arraystretch}{1.1}
\begin{tabular}{@{}ll ccc>{\columncolor{gray!12}}c@{}}
\toprule
\textbf{Method} & \textbf{Config} & \textbf{MT-Bench} & \textbf{Humaneval} & \textbf{GSM8K} & \textbf{Avg} \\
\midrule
Full Vocab & -- & 1.18 \Std{$\pm$ 0.01} & 1.58 \Std{$\pm$ 0.01} & 1.63 \Std{$\pm$ 0.01} & 1.46 \\
\addlinespace[2pt]

\multirow{2}{*}{FR-Spec} & $V_{\mathrm{tr}}=64$K & 1.18 \Std{$\pm$ 0.01} & 1.62 \Std{$\pm$ 0.01} & 1.63 \Std{$\pm$ 0.02} & 1.48 \\
 & $V_{\mathrm{tr}}=32$K & 1.13 \Std{$\pm$ 0.01} & 1.50 \Std{$\pm$ 0.01} & 1.49 \Std{$\pm$ 0.01} & 1.37 \\
\addlinespace[2pt]

\multirow{2}{*}{VocabTrim} & $V_{\mathrm{tr}}=64$K & 1.22 \Std{$\pm$ 0.02} & 1.67 \Std{$\pm$ 0.01} & 1.69 \Std{$\pm$ 0.01} & 1.53 \\
 & $V_{\mathrm{tr}}=32$K & 1.22 \Std{$\pm$ 0.02} & 1.66 \Std{$\pm$ 0.01} & 1.69 \Std{$\pm$ 0.02} & 1.52 \\
\addlinespace[2pt]

BCL & $V_{\mathrm{tr}}=66.9$K & 1.19 \Std{$\pm$ 0.02} & 1.57 \Std{$\pm$ 0.01} & 1.65 \Std{$\pm$ 0.02} & 1.47 \\
\addlinespace[2pt]

\multirow{2}{*}{VocabTrim-T} & $V_{\mathrm{tr}}=64$K & 1.22 \Std{$\pm$ 0.01} & 1.65 \Std{$\pm$ 0.02} & 1.70 \Std{$\pm$ 0.01} & 1.52 \\
 & $V_{\mathrm{tr}}=32$K & 1.21 \Std{$\pm$ 0.02} & 1.63 \Std{$\pm$ 0.02} & 1.69 \Std{$\pm$ 0.02} & 1.51 \\
\addlinespace[2pt]

SpecVocab & $r=d/8$ & {\bf 1.26} \Std{$\pm$ 0.01} & 1.67 \Std{$\pm$ 0.02} & 1.72 \Std{$\pm$ 0.01} & 1.55 \\
\midrule
\multirow{3}{*}{SlimSpec} & $r=d/4$ & 1.25 \Std{$\pm$ 0.01} & {\bf 1.73} \Std{$\pm$ 0.04} & 1.76 \Std{$\pm$ 0.02} & {\bf 1.58} \\
 & $r=d/8$ & {\bf 1.26} \Std{$\pm$ 0.01} & 1.72 \Std{$\pm$ 0.02} & {\bf 1.77} \Std{$\pm$ 0.01} & {\bf 1.58} \\
 & $r=d/16$ & 1.21 \Std{$\pm$ 0.01} & 1.71 \Std{$\pm$ 0.01} & 1.73 \Std{$\pm$ 0.01} & 1.55 \\
\bottomrule
\end{tabular}
\end{table}

\begin{table}[htbp]
\centering
\caption{Average acceptance length $\tau$ for Qwen3-30B-A3B, temperature $=1$, batch size $=1$.}
\label{tab:qwen3-30b-t1-bs1-tau}

\setlength{\tabcolsep}{9pt}
\renewcommand{\arraystretch}{1.1}
\begin{tabular}{@{}ll ccc>{\columncolor{gray!12}}c@{}}
\toprule
\textbf{Method} & \textbf{Config} & \textbf{MT-Bench} & \textbf{Humaneval} & \textbf{GSM8K} & \textbf{Avg} \\
\midrule
Full Vocab & -- & 2.69 \Std{$\pm$ 0.03} & 3.76 \Std{$\pm$ 0.04} & 3.94 \Std{$\pm$ 0.03} & 3.46 \\
\addlinespace[2pt]

\multirow{2}{*}{FR-Spec} & $V_{\mathrm{tr}}=64$K & 2.60 \Std{$\pm$ 0.02} & 3.69 \Std{$\pm$ 0.02} & 3.80 \Std{$\pm$ 0.03} & 3.36 \\
 & $V_{\mathrm{tr}}=32$K & 2.44 \Std{$\pm$ 0.02} & 3.34 \Std{$\pm$ 0.02} & 3.39 \Std{$\pm$ 0.01} & 3.06 \\
\addlinespace[2pt]

\multirow{2}{*}{VocabTrim} & $V_{\mathrm{tr}}=64$K & 2.70 \Std{$\pm$ 0.02} & 3.75 \Std{$\pm$ 0.03} & 3.93 \Std{$\pm$ 0.02} & 3.46 \\
 & $V_{\mathrm{tr}}=32$K & 2.66 \Std{$\pm$ 0.03} & 3.70 \Std{$\pm$ 0.03} & 3.86 \Std{$\pm$ 0.03} & 3.41 \\
\addlinespace[2pt]

BCL & $V_{\mathrm{tr}}=66.9$K & 2.55 \Std{$\pm$ 0.02} & 3.48 \Std{$\pm$ 0.03} & 3.76 \Std{$\pm$ 0.01} & 3.26 \\
\addlinespace[2pt]

\multirow{2}{*}{VocabTrim-T} & $V_{\mathrm{tr}}=64$K & 2.67 \Std{$\pm$ 0.02} & 3.72 \Std{$\pm$ 0.05} & 3.92 \Std{$\pm$ 0.02} & 3.44 \\
 & $V_{\mathrm{tr}}=32$K & 2.64 \Std{$\pm$ 0.02} & 3.62 \Std{$\pm$ 0.04} & 3.86 \Std{$\pm$ 0.01} & 3.37 \\
\addlinespace[2pt]

SpecVocab & $r=d/8$ & 2.69 \Std{$\pm$ 0.01} & 3.76 \Std{$\pm$ 0.04} & 3.97 \Std{$\pm$ 0.03} & 3.47 \\
\midrule
\multirow{3}{*}{SlimSpec} & $r=d/4$ & 2.65 \Std{$\pm$ 0.02} & 3.83 \Std{$\pm$ 0.05} & 3.99 \Std{$\pm$ 0.02} & 3.49 \\
 & $r=d/8$ & 2.61 \Std{$\pm$ 0.02} & 3.75 \Std{$\pm$ 0.03} & 3.96 \Std{$\pm$ 0.02} & 3.44 \\
 & $r=d/16$ & 2.53 \Std{$\pm$ 0.02} & 3.71 \Std{$\pm$ 0.03} & 3.87 \Std{$\pm$ 0.01} & 3.37 \\
\bottomrule
\end{tabular}
\end{table}

\clearpage
\begin{table}[htbp]
\centering
\caption{Speedup for Qwen3-30B-A3B, temperature $=1$, batch size $=64$.}
\label{tab:qwen3-30b-t1-bs64-speedup}

\setlength{\tabcolsep}{9pt}
\renewcommand{\arraystretch}{1.1}
\begin{tabular}{@{}ll ccc>{\columncolor{gray!12}}c@{}}
\toprule
\textbf{Method} & \textbf{Config} & \textbf{MT-Bench} & \textbf{Humaneval} & \textbf{GSM8K} & \textbf{Avg} \\
\midrule
Full Vocab & -- & 1.11 \Std{$\pm$ 0.03} & 1.21 \Std{$\pm$ 0.04} & 1.32 \Std{$\pm$ 0.01} & 1.21 \\
\addlinespace[2pt]

\multirow{2}{*}{FR-Spec} & $V_{\mathrm{tr}}=64$K & 1.19 \Std{$\pm$ 0.02} & 1.23 \Std{$\pm$ 0.04} & 1.36 \Std{$\pm$ 0.01} & 1.26 \\
 & $V_{\mathrm{tr}}=32$K & 1.11 \Std{$\pm$ 0.07} & 1.20 \Std{$\pm$ 0.05} & 1.28 \Std{$\pm$ 0.03} & 1.20 \\
\addlinespace[2pt]

\multirow{2}{*}{VocabTrim} & $V_{\mathrm{tr}}=64$K & 1.19 \Std{$\pm$ 0.02} & 1.23 \Std{$\pm$ 0.03} & 1.37 \Std{$\pm$ 0.03} & 1.26 \\
 & $V_{\mathrm{tr}}=32$K & 1.18 \Std{$\pm$ 0.01} & 1.27 \Std{$\pm$ 0.04} & 1.36 \Std{$\pm$ 0.02} & 1.27 \\
\addlinespace[2pt]

BCL & $V_{\mathrm{tr}}=66.9$K & {\bf 1.20} \Std{$\pm$ 0.02} & 1.26 \Std{$\pm$ 0.02} & 1.35 \Std{$\pm$ 0.02} & 1.27 \\
\addlinespace[2pt]

\multirow{2}{*}{VocabTrim-T} & $V_{\mathrm{tr}}=64$K & 1.19 \Std{$\pm$ 0.03} & 1.27 \Std{$\pm$ 0.04} & 1.38 \Std{$\pm$ 0.01} & 1.28 \\
 & $V_{\mathrm{tr}}=32$K & 1.18 \Std{$\pm$ 0.02} & 1.26 \Std{$\pm$ 0.02} & 1.35 \Std{$\pm$ 0.05} & 1.26 \\
\addlinespace[2pt]

SpecVocab & $r=d/8$ & 1.09 \Std{$\pm$ 0.17} & 1.24 \Std{$\pm$ 0.04} & 1.32 \Std{$\pm$ 0.10} & 1.21 \\
\midrule
\multirow{3}{*}{SlimSpec} & $r=d/4$ & 1.18 \Std{$\pm$ 0.04} & 1.19 \Std{$\pm$ 0.20} & 1.35 \Std{$\pm$ 0.10} & 1.24 \\
 & $r=d/8$ & 1.19 \Std{$\pm$ 0.02} & 1.28 \Std{$\pm$ 0.02} & {\bf 1.39} \Std{$\pm$ 0.04} & {\bf 1.29} \\
 & $r=d/16$ & 1.18 \Std{$\pm$ 0.03} & {\bf 1.29} \Std{$\pm$ 0.02} & 1.34 \Std{$\pm$ 0.09} & 1.27 \\
\bottomrule
\end{tabular}
\end{table}

\begin{table}[htbp]
\centering
\caption{Average acceptance length $\tau$ for Qwen3-30B-A3B, temperature $=1$, batch size $=64$.}
\label{tab:qwen3-30b-t1-bs64-tau}

\setlength{\tabcolsep}{9pt}
\renewcommand{\arraystretch}{1.1}
\begin{tabular}{@{}ll ccc>{\columncolor{gray!12}}c@{}}
\toprule
\textbf{Method} & \textbf{Config} & \textbf{MT-Bench} & \textbf{Humaneval} & \textbf{GSM8K} & \textbf{Avg} \\
\midrule
Full Vocab & -- & 2.66 \Std{$\pm$ 0.00} & 3.70 \Std{$\pm$ 0.07} & 3.88 \Std{$\pm$ 0.04} & 3.41 \\
\addlinespace[2pt]

\multirow{2}{*}{FR-Spec} & $V_{\mathrm{tr}}=64$K & 2.58 \Std{$\pm$ 0.02} & 3.60 \Std{$\pm$ 0.06} & 3.72 \Std{$\pm$ 0.06} & 3.30 \\
 & $V_{\mathrm{tr}}=32$K & 2.43 \Std{$\pm$ 0.02} & 3.27 \Std{$\pm$ 0.02} & 3.37 \Std{$\pm$ 0.03} & 3.02 \\
\addlinespace[2pt]

\multirow{2}{*}{VocabTrim} & $V_{\mathrm{tr}}=64$K & 2.66 \Std{$\pm$ 0.03} & 3.67 \Std{$\pm$ 0.08} & 3.87 \Std{$\pm$ 0.03} & 3.40 \\
 & $V_{\mathrm{tr}}=32$K & 2.63 \Std{$\pm$ 0.02} & 3.61 \Std{$\pm$ 0.07} & 3.81 \Std{$\pm$ 0.05} & 3.35 \\
\addlinespace[2pt]

BCL & $V_{\mathrm{tr}}=66.9$K & 2.57 \Std{$\pm$ 0.02} & 3.43 \Std{$\pm$ 0.04} & 3.75 \Std{$\pm$ 0.02} & 3.25 \\
\addlinespace[2pt]

\multirow{2}{*}{VocabTrim-T} & $V_{\mathrm{tr}}=64$K & 2.67 \Std{$\pm$ 0.02} & 3.69 \Std{$\pm$ 0.03} & 3.92 \Std{$\pm$ 0.02} & 3.43 \\
 & $V_{\mathrm{tr}}=32$K & 2.62 \Std{$\pm$ 0.02} & 3.61 \Std{$\pm$ 0.03} & 3.79 \Std{$\pm$ 0.01} & 3.34 \\
\addlinespace[2pt]

SpecVocab & $r=d/8$ & 2.67 \Std{$\pm$ 0.02} & 3.74 \Std{$\pm$ 0.02} & 3.95 \Std{$\pm$ 0.04} & 3.45 \\
\midrule
\multirow{3}{*}{SlimSpec} & $r=d/4$ & 2.67 \Std{$\pm$ 0.03} & 3.82 \Std{$\pm$ 0.04} & 3.97 \Std{$\pm$ 0.03} & 3.49 \\
 & $r=d/8$ & 2.59 \Std{$\pm$ 0.01} & 3.72 \Std{$\pm$ 0.01} & 3.94 \Std{$\pm$ 0.02} & 3.42 \\
 & $r=d/16$ & 2.52 \Std{$\pm$ 0.02} & 3.68 \Std{$\pm$ 0.03} & 3.85 \Std{$\pm$ 0.03} & 3.35 \\
\bottomrule
\end{tabular}
\end{table}




\end{document}